%% file: iclr2027_conference.tex
\documentclass{article} 
\usepackage{iclr2027_conference,times}

\input{math_commands.tex}

\usepackage{hyperref}
\usepackage{url}

\usepackage{amsmath} 
\usepackage{subcaption}
\usepackage{graphicx}
\usepackage{array}

\usepackage[utf8]{inputenc} 
\usepackage[T1]{fontenc}    
\usepackage{hyperref}       
\usepackage{url}            
\usepackage{booktabs}       
\usepackage{amsfonts}       
\usepackage{nicefrac}       
\usepackage{microtype}      
\usepackage{xcolor}         
\usepackage{multirow}
\usepackage{float}
\usepackage{comment}

\usepackage{pifont}
\usepackage{placeins}

\usepackage[table]{xcolor}

\definecolor{modelgray}{RGB}{235,235,235}
\definecolor{oursgreen}{RGB}{225,242,220}
\usepackage{xcolor}

\title{SIFT: Enhancing Time Series Foundation Models via Semantic Invariance and Structural Fidelity Fine-Tuning}

\author{
Yi Tang$^{1}$, Tengxue Zhang$^{1}$, Yang Shu$^{1}$, Chenjuan Guo$^{1}$,
Chenchen Sun$^{2}$, Yisheng An$^{2}$ \\
\small
$^{1}$East China Normal University, Shanghai, China \\
$^{2}$Chang'an University, Xi'an, China \\
}

\iclrfinalcopy
\begin{document}

\maketitle

\begin{abstract}
Time Series Foundation Models (TSFMs) have achieved remarkable zero-shot performance through extensive pre-training on massive time series datasets. 
Nevertheless, due to the low-dimensional properties and diverse structural patterns of time series data, performing naive fine-tuning on TSFMs often leads to overfitting and falling into the mean-prediction trap. 
To address these challenges, we propose \textbf{SIFT}, a robust adaptation method that enhances time series foundation models by preserving \textbf{S}emantic \textbf{I}nvariance and structural \textbf{F}idelity throughout the fine-\textbf{T}uning process.
We employ semantic-invariant adversarial augmentation, which utilizes semantic spectrum decomposition to partition the semantic space and then generates perturbations within the non-core semantic subspace to bolster the model’s robustness against these perturbations, mitigating overfitting. 
We implement a component-based structural fidelity enhancement, which facilitates component-wise mixup and imposes a reconstruction objective to improve the model's ability to preserve structural fidelity, alleviating the mean-prediction trap. 
Extensive experiments on representative TSFMs covering 10 real-world datasets demonstrate that \textbf{SIFT} can significantly enhance the performance of TSFMs.
\end{abstract}

\section{Introduction}
Recently, time series foundation models (TSFMs) have emerged as a promising paradigm for time series analysis~\citep{DUET-2025, APN-2026, crossad-2025}. By pre-training on massive time series datasets, TSFMs acquire a broad understanding of temporal patterns, enabling them to deliver impressive zero-shot forecasting performance~\citep {Moirai-2024, Moment-2024, unitime-2024}. 
However, real-world time series frequently exhibit subtle temporal characteristics and domain-dependent variations that are difficult to fully capture during pre-training. As a result, such zero-shot forecasting often offers suboptimal results in practical applications. 
To unlock the full potential of TSFMs, a dedicated fine-tuning method is essential to adapt the model for the nuanced and intricate temporal patterns inherent in specialized downstream scenarios.


Currently, a broad range of parameter-efficient fine-tuning (PEFT) methods have been proposed, including LoRA~\citep{lora-2022}, AdaLoRA~\citep{adalora-2023}, Prefix Tuning~\citep{prefix-2021}, and Linear Probing~\citep{linear-2022}. These methods are primarily developed in NLP and CV and have been widely adapted to diverse downstream tasks. More recently, TSFM-specific approaches such as MSFT~\citep{MSFT-2025} have been introduced to better accommodate time series by explicitly integrating multi-scale\citep{MICN-2023} modeling into the fine-tuning process. However, existing methods still struggle to preserve semantic consistency and fine-grained structural information in time series data during fine-tuning.
Several critical challenges therefore remain unresolved.

\begin{figure}[t]
  \centering

  \includegraphics[width=\textwidth]{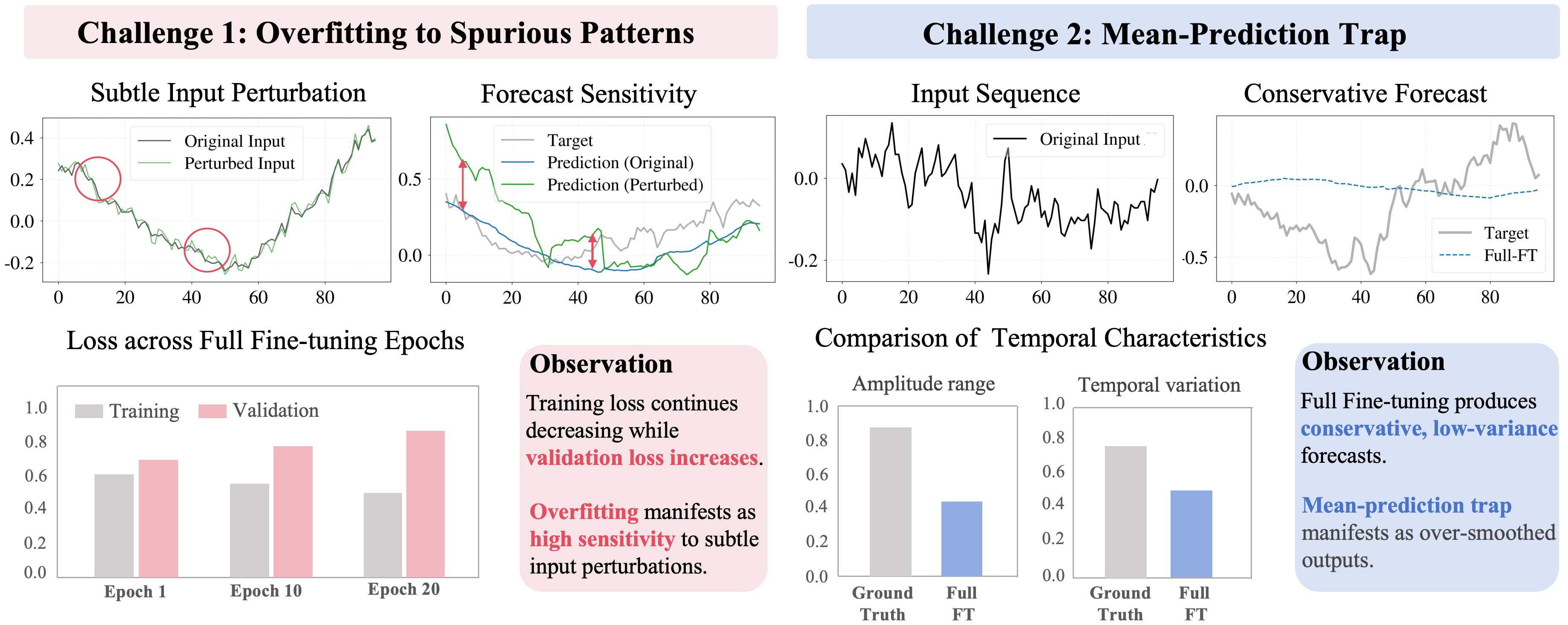}
  \caption{Two primary bottlenecks in TSFM fine-tuning. (1) Overfitting to intrinsic low-dimensional structures results in high sensitivity to subtle input perturbations. (2) Downstream scenarios that deviate significantly from pre-training patterns lead to over-smoothed outputs and structural absence.}
  \label{fig:challenges}
\end{figure}

\textbf{Challenge 1: TSFMs are prone to overfitting when adapting to low-dimensional time series.}
TSFMs are typically pre-trained with large-scale architectures to learn transferable representations across diverse time-series domains.
However, downstream tasks often involve time series exhibiting intrinsic low-dimensional structures~\citep{Timebase-2025, low-2025}, where core semantics, such as long-term trends and seasonal patterns, are concentrated in a sparse subspace.
This mismatch between the high-capacity pre-trained model and the low-dimensional downstream data increases the risk of overfitting to spurious patterns or non-transferable variations during fine-tuning.
Such overfitting often manifests as high sensitivity to input perturbations.
As shown in Figure~\ref{fig:challenges}, even subtle perturbations in the input can induce disproportionately large deviations in the output space, indicating poor robustness to local fluctuations or minor distribution shifts introduced during fine-tuning.
Consequently, this fragility hinders effective transfer of pre-trained temporal knowledge and semantic representations~\citep{pre-train-2024} to downstream tasks~\citep{BSS-2019}.

\textbf{Challenge 2: TSFMs are susceptible to the mean-prediction trap under domain divergence.} 
While TSFMs internalize transferable temporal priors through massive pre-training, their generalization capability is often compromised when downstream datasets deviate substantially from those pre-training distributions. 
Modern TSFMs employ diverse forecasting objectives, including pointwise regression losses, likelihood-based distribution modeling, and quantile regression.
Despite these different formulations, they primarily optimize numerical prediction accuracy or predictive distribution quality during fine-tuning.
Under domain divergence, this lack of explicit structural preservation may lead the model to underrepresent fine-grained temporal dynamics, resulting in conservative, low-variance forecasts—a phenomenon referred to as the mean-prediction trap~\citep{Forecast-Collapse}.
As illustrated in Figure~\ref{fig:challenges}, this manifests as pronounced over-smoothing~\citep{oversmooth-2025}, suppressed peaks, and a tendency to regress toward the average trajectory. Such behavior undermines the structural fidelity of forecasts, limiting the model's ability to capture high-amplitude temporal variations when downstream patterns diverge from the pre-trained priors.

To address \textbf{Challenge 1}, we propose a \textbf{semantic-invariant adversarial augmentation} strategy to improve model robustness against perturbations that do not alter core semantics, reducing overfitting to spurious patterns and non-transferable variations.
There are two key components in this strategy: (1) \textit{semantic spectrum decomposition}, which partitions the semantic space into subspaces to identify the perturbation space for the adversarial attacks; and (2) \textit{adversarial perturbation augmentation}, which first produces input perturbations that maximally expose the model’s sensitivity while preserving the core semantics. 
During fine-tuning, we minimize the discrepancy between the model’s predictions on original and perturbed inputs, compelling the model to prioritize core semantic features over non-core variations. 
This strategy enables the model to capture core semantic features effectively, while improving robustness by reducing sensitivity to input perturbations.

To address \textbf{Challenge 2}, we propose a \textbf{component-based structural fidelity enhancement} framework to mitigate the mean-prediction trap by guiding the model to reconstruct high-fidelity structures. 
To prevent the model from collapsing into conservative, low-variance forecasts, we adopt \textit{structural component mixup} to synthesize structurally diverse hard samples, which first prioritizes high-volatility samples then decomposes these samples into latent trend and seasonal components to facilitate component-wise mixup.
Furthermore, we impose \textit{component-wise fidelity reconstruction objective} to provide fine-grained structural supervision, encouraging the model to preserve structural fidelity. Therefore, the model acquires the capacity to track volatile fluctuations, alleviating the over-smoothing and the mean-prediction phenomenon.

Overall, we propose SIFT, a fine-tuning method for time series foundation models that preserves semantic invariance and structural fidelity, enabling stable adaptation to diverse downstream time series forecasting tasks. Our primary contributions are summarized as follows:

\begin{itemize}
 \item We propose a semantic-invariant adversarial augmentation strategy to reduce overfitting to spurious patterns and prioritize core semantics, ensuring robust fine-tuning of TSFMs.
 \item We develop a component-based structural fidelity enhancement to bolster model adaptability to distributional shifts, mitigating over-smoothing and the mean-prediction trap.
  \item Extensive experiments on 10 real-world downstream datasets using 4 representative pre-trained models demonstrate that our method outperforms existing baselines, achieving state-of-the-art performance in the fine-tuning of TSFMs for time series forecasting.
\end{itemize}

\section{Related Work}

\textbf{Time Series Foundation Model.}
Recent years have witnessed a surge in time series foundation models (TSFMs). Existing architectures can be broadly categorized into three primary architectures:
(1) Encoder-only paradigms: MOIRAI ~\citep{Moirai-2024} unifies diverse multivariate inputs through sequence flattening, while MOMENT ~\citep{Moment-2024} and UniTS~\citep{Units-2024} utilize masked reconstruction and task-specific tokenization to capture high-level features across varying domains.
(2) Decoder-only architectures: Inspired by Large Language Models, frameworks like TimesFM ~\citep{TimesFM-2024} and Timer ~\citep{Timer-2024} treat forecasting as an autoregressive process. 
(3) Encoder-decoder and generative variants: This category focuses on integrating structural or domain-specific priors into the generative process. Notable examples include TTM ~\citep{TTM-2024}, which leverages multi-resolution sampling for cross-channel patterns, and ROSE ~\citep{Rose-2024}, which emphasizes frequency-domain decomposition. Furthermore, Sundial ~\citep{Sundial-2025} introduces a generative Transformer framework for modeling continuous temporal flows. However, while TSFMs have achieved remarkable breakthroughs in zero-shot scenarios, how to effectively perform fine-tuning on downstream tasks remains an under-explored yet critical problem. 

\textbf{Fine-tuning for TSFMs.}
Methods for fine-tuning TSFMs have largely inherited Parameter-Efficient Fine-Tuning (PEFT) techniques from NLP and CV, which can be broadly categorized into three main paradigms: (1) Selective fine-tuning, such as Linear Probing \citep{linear-2022}, which freezes the backbone and updates only the task-specific output head; (2) Additive fine-tuning, notably Prefix Tuning \citep{prefix-2021}, which introduces trainable continuous vectors to hidden states to guide the model’s representations; and (3) Reparameterization-based fine-tuning, including LoRA \citep{lora-2022} and its variant AdaLoRA \citep{adalora-2023}, which adapt internal weights through low-rank decomposition and importance-based budget allocation. While efficient, these general-purpose strategies are often insufficient to handle the unique non-stationarity of temporal signals. Consequently, recent research has pivoted toward paradigms tailored specifically for the temporal domain, notably MSFT ~\citep{MSFT-2025}. By adopting a causal perspective, MSFT models multiple temporal scales~\citep{scale-2021, scale-2022} to leverage the forecasting potential of backbones. Despite these advances, existing paradigms often still overlook the overfitting and over-smoothing phenomenon during practical downstream model adaptation.

\section{Methodology}
\label{others}
\textbf{Problem Formulation.} In multivariate time series forecasting, the goal is to predict future values $Y \in \mathbb{R}^{H \times C}$ based on historical observations $X \in \mathbb{R}^{L \times C}$, where $L$, $H$, and $C$ denote the look-back length, forecasting horizon, and number of variates, respectively. In the context of fine-tuning, we leverage a time series foundation model $f(\cdot)$ with parameters $\theta$, which has been pre-trained on large-scale source datasets to capture general temporal patterns. Given a target dataset $\mathcal{D} = \{(X, Y)\}$ from a specific downstream domain, our objective is to fine-tune $\theta$ on the target domain, optimizing the model to minimize forecasting error and achieve superior performance on downstream tasks.

\begin{figure}[t] 
  \centering
  \includegraphics[width=1.0\linewidth]{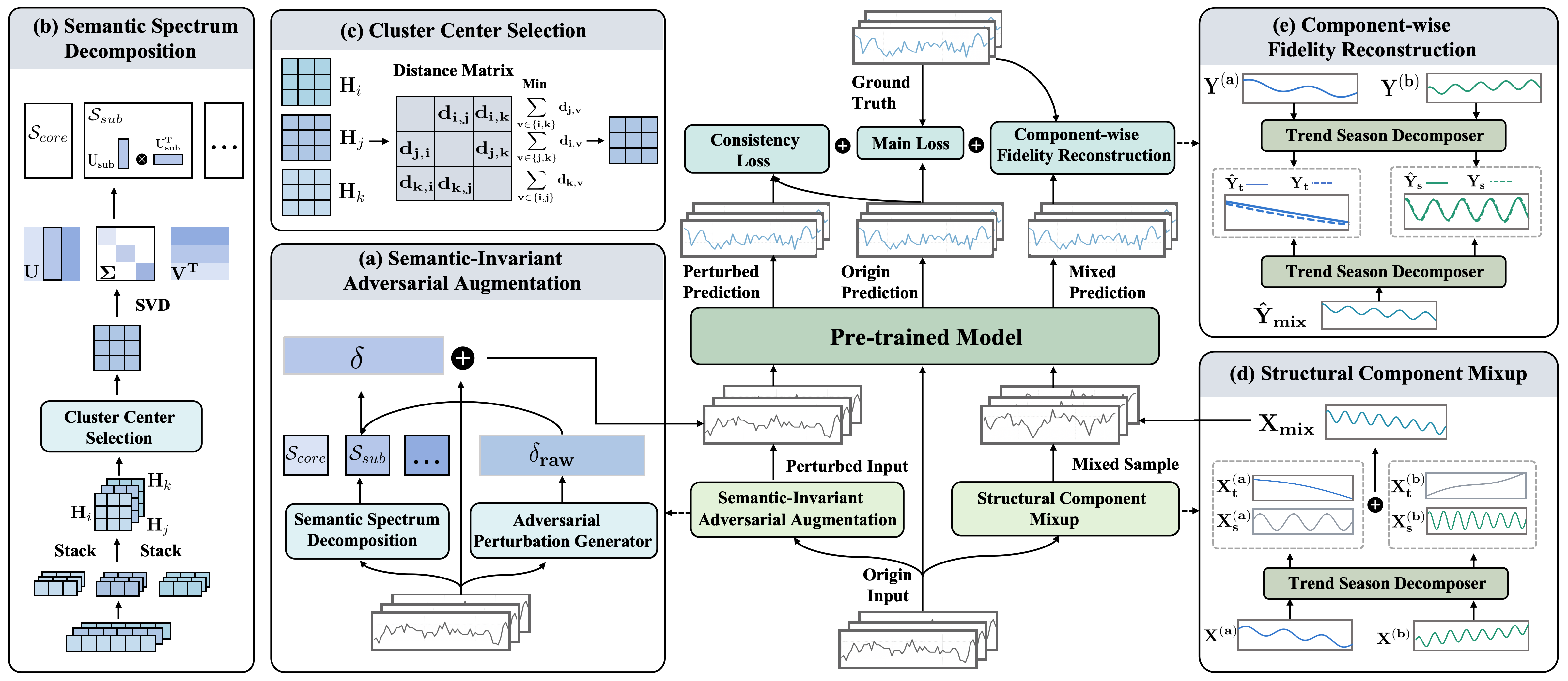} 
  \caption{The framework of SIFT, comprising two main modules: semantic-invariant adversarial augmentation (SIAA) to mitigate overfitting and component-based structural fidelity enhancement (CSFE) to preserve temporal structure through component mixup and reconstruction.}
  \label{fig:framework}
\end{figure}

We propose SIFT, a fine-tuning method for time series foundation models that preserves semantic invariance and structural fidelity (Figure~\ref{fig:framework}) :
(1) The \textbf{semantic-invariant adversarial augmentation} strategy is designed to bolster model robustness and mitigate overfitting by enhancing the model's ability to capture core semantic features. A semantic spectrum decomposition is employed to partition the semantic space. Then, an adversarial perturbation generator produces perturbations to maximize TSFM's prediction divergence between the original and perturbed inputs, while a subspace projection using the predefined partitioning ensures these perturbations are confined to the subordinate semantic space to preserve semantic invariance. During fine-tuning, the discrepancy between the model's predictions for original and perturbed samples is minimized. 
(2) The \textbf{component-based structural fidelity enhancement} is proposed to mitigate the mean-prediction trap by compelling the model to reconstruct structural details. We first identify training samples exhibiting high temporal volatility. These samples perform trend-seasonal decomposition, followed by a structural component mixup that recombines their decomposed components to synthesize structurally diverse hard examples. Finally, we use a component-wise fidelity reconstruction loss to align the decomposed trend and seasonal components of the outputs with those of the ground-truth, preserving the sharp structural fidelity.

\subsection{Semantic-Invariant Adversarial Augmentation}
To enhance the robustness of TSFMs during fine-tuning and mitigate overfitting, we propose a semantic-invariant adversarial augmentation strategy. The key idea is to generate perturbations within a semantically constrained subspace, preserving core semantic features while amplifying non-core variations to expose the model's sensitivity. During fine-tuning, we enforce prediction consistency under such subspace-constrained perturbations, encouraging the model to focus on the stable core semantics rather than spurious patterns or non-transferable variations.

\textbf{Semantic Spectrum Decomposition.}
\label{sec:SSD}
Instead of introducing perturbations directly in the raw signal space that risks degrading essential semantic structures, we first perform a semantic decomposition to disentangle the underlying structured components of the input. For simplicity, we describe this process using a single time series channel. 
Given a time series input $X \in \mathbb{R}^{L}$, we define a trajectory embedding operator $\Psi$ that maps 
1D time series $X$ into a trajectory matrix $\mathbf{H}$~\citep{SSA-2002}:
\begin{equation}
\mathbf{H} = \Psi(X) =
\begin{bmatrix}
X(1) & X(1+S) & \cdots & X(1+(K-1)S) \\
X(2) & X(2+S) & \cdots & X(2+(K-1)S) \\
\vdots & \vdots & \ddots & \vdots \\
X(W) & X(W+S) & \cdots & X(W+(K-1)S) 
\end{bmatrix}\in \mathbb{R}^{W \times K}
\end{equation}
where $W$ and $S$ denote the window width and stride, respectively, and the column dimension $K$ is given by $K = \lfloor \frac{L-W}{S} \rfloor + 1$. This transformation converts the original temporal signal into a matrix representation that explicitly exposes local temporal dependencies and structural patterns.

Inspired by prior spectral analyses~\citep{BSS-2019}, we perform singular value decomposition (SVD) on the trajectory matrix to characterize its internal spectral structure:
\begin{equation}
\mathbf{H} = \mathbf{U}\boldsymbol{\Sigma}\mathbf{V}^{T},
\end{equation}
where $\mathbf{U} \in \mathbb{R}^{W \times d}$ and $\mathbf{V} \in \mathbb{R}^{K \times d}$ are orthonormal matrices containing the left and right singular vectors, $\boldsymbol{\Sigma} = \operatorname{diag}(\sigma_1, \dots, \sigma_d)$ with $\sigma_1 \ge \cdots \ge \sigma_d \ge 0$, and $d = \min(W, K)$. To obtain a semantic partition of the spectrum, we define two rank thresholds, $r_1 = \lfloor d \cdot \rho_1 \rfloor$ and $r_2 = \lfloor d \cdot \rho_2 \rfloor$, where scaling factors $0 < \rho_1 < \rho_2 \le 1$. These thresholds divide the left singular vectors (indexed by $i = 1, \dots, d$) into three groups, 
$\mathbf{U}_{\text{core}} = [\mathbf{u}_1, \dots, \mathbf{u}_{r_1}]$,
$\mathbf{U}_{\text{sub}} = [\mathbf{u}_{r_1+1}, \dots, \mathbf{u}_{r_2}]$, and
$\mathbf{U}_{\text{noise}} = [\mathbf{u}_{r_2+1}, \dots, \mathbf{u}_d]$, spanning the core subspace ($\mathcal{S}_{\text{core}}$), the subordinate subspace ($\mathcal{S}_{\text{sub}}$), and the noise subspace ($\mathcal{S}_{\text{noise}}$). 

According to the \textit{Eckart-Young-Mirsky theorem}~\citep{EYM-1936}, the optimal rank-$r_1$ approximation of $\mathbf{H}$, which we denote as $\mathbf{H}_{\text{core}}$, is given by truncating the SVD:
\begin{equation}
\mathbf{H}_{\text{core}} = \arg\min_{\text{rank}(\mathbf{H}')=r_1} \|\mathbf{H} - \mathbf{H}'\|_F = \sum_{i=1}^{r_1} \sigma_i \mathbf{u}_i \mathbf{v}_i^T,
\end{equation}
with a minimum Frobenius norm error of $\sqrt{\sum_{i=r_1+1}^{d} \sigma_i^2}$. This indicates that the subspace $\mathcal{S}_{\text{core}}$ captures the dominant semantic structures, which typically correspond to long-term trends and seasonal patterns. The noise boundary $r_2$ is inspired by the \textit{Marchenko-Pastur (MP) law}~\citep{MP-1967}, which provides an empirical approximation for separating signal and noise in spectral analysis.
The singular values associated with noise are expected to concentrate below the threshold:
\begin{equation}
\lambda_{+} = \hat{\sigma}_{\epsilon} \left(1 + \sqrt{\frac{W}{K}}\right),
\end{equation}
where $\hat{\sigma}_{\epsilon}$ is estimated from the tail of the spectrum, and $W$ and $K$ denote the dimensions of the matrix $\mathbf{H}$.
We characterize $r_2$ as the largest index satisfying $\sigma_{r_2} > \lambda_{+}$. The subspace spanned by the singular vectors with indices greater than $r_2$, denoted $\mathcal{S}_{\text{noise}}$, consists of stochastic fluctuations.

The singular components indexed between \(r_1+1\) and \(r_2\) span the subordinate semantic subspace \(\mathcal{S}_{\text{sub}}\).
This subspace captures components that are neither core temporal semantics encoded in \(\mathcal{S}_{\text{core}}\) nor stochastic fluctuations contained in \(\mathcal{S}_{\text{noise}}\).
Such components are more likely to encode spurious patterns or non-transferable variations, making the model particularly prone to overfitting to them. 
We therefore use this subspace as the primary domain for adversarial perturbation. 
For multivariate time series, we first cluster channels according to their temporal similarity and use the cluster centers as representative sequences. We then perform spectral stratification on each cluster center.

\textbf{Adversarial Perturbation Augmentation.} 
To enhance the robustness of the TSFM during fine-tuning, we design a two-stage adversarial framework. First, we train a generator to produce semantic-invariant perturbations that preserve the core semantic subspace while substantially disrupting the model's predictions. After the generator is trained, it is fixed and used to fine-tune the TSFM to maintain invariant predictions under such perturbations.

\textit{Stage 1: Generator Training.} In this stage, the foundation model $f$ is frozen. Our goal is to train a perturbation generator \(PG\) to produce effective adversarial perturbations that maximally disrupt the model's predictions while preserving temporal semantics.

The perturbation is generated as follows. Given an input $X$, the generator $PG$ first produces a raw perturbation $\delta_\text{raw} = PG(X)$, where $PG$ is a lightweight convolutional neural network~\citep{CNN-2020, CNN-1998}. To ensure semantic invariance, we project $\delta_\text{raw}$ into the subordinate semantic subspace using the projection matrix $\mathbf{P}_\text{sub} = \mathbf{U}_\text{sub}\mathbf{U}_\text{sub}^{\top}$, and the resulting semantic-invariant perturbation $\delta_{\text{adv}}$ is reconstructed as:
\begin{equation}
\delta_{\text{adv}} = \Psi^{-1}\left( \mathbf{P}_\text{sub} \cdot \Psi(\delta_\text{raw}) \right)
\end{equation}
where $\Psi(\cdot)$ denotes the trajectory operator that maps a time series into its trajectory matrix representation, and $\Psi^{-1}(\cdot)$ is the anti-diagonal averaging operator that reconstructs a time series from a trajectory matrix by averaging along its anti-diagonals. 
The final perturbed input is then $X_{\text{adv}} = X + \delta_{\text{adv}}$.
To constrain leakage into the core semantic space, we define
$\delta_{\text{core}} = \Psi^{-1}(\mathbf{U}_{\text{core}}\mathbf{U}_{\text{core}}^{T}\Psi(\delta_{\text{raw}}))$,
which represents the projection of the raw perturbation into the core semantic subspace.


With $X_{\text{adv}}$ and $\delta_{\text{core}}$ defined above, $PG$ is optimized to minimize the following adversarial loss:
\begin{equation}
\min_{\theta_{PG}} \mathcal{L}_{PG}
=
-
\underbrace{
\left\|f(X_{\text{adv}})-f(X)\right\|_{2}^{2}
}_{\text{Adversarial Disruption}}
+
\underbrace{
\left\|\delta_{\text{core}}\right\|_{2}^{2}
}_{\text{Core Leakage Penalty}}.
\end{equation}

\textit{Stage 2 Robust Fine-tuning.} Once the generator is trained, the generator $PG$ is fixed to serve as a dedicated module for augmentation, producing semantic-invariant perturbed inputs. During fine-tuning, these inputs are used to train the model, enforcing prediction consistency under such perturbations. To achieve this, we minimize the following consistency loss:
\begin{equation}
\mathcal{L}_{\text{consis}} = \left\|f(X_{\text{adv}}) - f(X)\right\|_{2}^{2}
\end{equation}
This encourages the model to focus on core semantic features rather than spurious patterns, thereby maintaining stable predictions, mitigating overfitting, and enhancing robustness. 

\subsection{Component-based Structural Fidelity Enhancement}

\textbf{Structural Component Mixup.} 
To mitigate the mean-prediction failure mode where models produce overly smooth outputs, we first prioritize samples with high temporal volatility as they contain richer temporal dynamics beyond smooth trends. 
The volatility of a sample $X \in \mathbb{R}^{L \times C}$ is characterized by the Total Variation (TV) score, which measures the cumulative magnitude of temporal shifts:
\begin{equation}
TV(X) = \frac{1}{C} \sum_{c=1}^{C} \sum_{t=1}^{L-1} |x_{c, t+1} - x_{c, t}|
\end{equation}
where $x_{c,t}$ denotes the observation value of the $c$-th channel at time step $t$. 
By ranking samples within each training batch according to their TV scores, we select the top-$k$ most volatile segments. Formally, we define the indices of these selected segments as the high-volatility set $I_{\text{top}}$, which serves as the foundation samples for the structural mixup. 

The mixed sample $X_{\text{mix}}$ is synthesized by cross-pairing the constituent components of an anchor sample $X^{(a)}$ and an arbitrary sample $X^{(b)}$, both sampled from $I_{\text{top}}$ ($a \neq b$). Using the Series Decomposition (SD) operation ~\citep{autoformer-2021}, we decouple them into their respective Trend ($X_t$) and Seasonal ($X_s$) components, denoted as $X_t, X_s = \operatorname{SD}(X)$. The mixed sample $X_{\text{mix}}$ is then synthesized by recombining these components:
\begin{equation}
X_{\text{mix}} = X_{t}^{(a)} + X_{s}^{(b)}
\end{equation}

\textbf{Component-wise Fidelity Reconstruction.}
To provide fine-grained structural supervision, we explicitly supervise that the model’s prediction for the mixed input remains faithful to the corresponding structural components of its original sources. 
Given the prediction $\hat{Y}_{\text{mix}} = f(X_{\text{mix}})$, we first extract its predicted trend and seasonal components:
\begin{equation}
{ \hat{Y}_{t}, \hat{Y}_{s} } = \operatorname{SD}(\hat{Y}_{\text{mix}})
\end{equation}
The structural reinforcement loss is then defined by aligning these predicted components with their corresponding cross-sample ground truths:
\begin{equation}
\mathcal{L}_t = \left\| \hat{Y}_{t} - Y_{t}^{(a)} \right\|_{1}, \quad
\mathcal{L}_s = \left\| \hat{Y}_{s} - Y_{s}^{(b)} \right\|_{2}^{2}
\end{equation}
We denote $Y^{(a)}_{t}$ and $Y^{(b)}_{s}$ as the ground-truth trend and seasonal components from source samples $a$ and $b$, respectively.
To balance the contributions of different structural components, we introduce an adaptive scaling factor $\alpha = \|X_s\| / \|X_t\|$, which normalizes the relative magnitudes of the seasonal and trend components. The integrated reconstruction loss is then formulated as:
\begin{equation}
\mathcal{L}_{\text{CFR}} = \mathcal{L}_s + \alpha \cdot \mathcal{L}_t
\end{equation}
By providing explicit supervision for individual structural attributes, the model is discouraged from defaulting to a conservative but inaccurate mean prediction. Instead, the separate objective on $\mathcal{L}_t$ and $\mathcal{L}_s$ forces the model to preserve the sharp structural fidelity of both long-term trends and periodic oscillations, effectively preventing the predictive blurring even under complex structural variations.

\subsection{Training and Inference}
\textit{Training Phase.} 
During training, we fine-tune the foundation model $f$ by minimizing a joint objective. The total loss $\mathcal{L}_{\text{total}}$ is a weighted combination of standard forecasting supervision, the proposed semantic consistency loss and the reconstruction loss:
\begin{equation}
\mathcal{L}_{\text{total}} = \left\| f(X) - Y \right\|_{2}^{2} + \lambda_{\text{consis}} \mathcal{L}_{\text{consis}} + \lambda_{\text{CFR}}\mathcal{L}_{\text{CFR}}
\end{equation}
The first term denotes the task-level forecasting loss, instantiated as MSE in our experiments.
The consistency loss $\mathcal{L}_{\text{consis}}$ encourages stable predictions under input perturbations, while the reconstruction loss $\mathcal{L}_{\text{CFR}}$ preserves the structural fidelity. Together, these regularizers prevent the foundation model from overfitting to spurious variations and mitigate the mean-prediction trap.

\textit{Inference Phase.} 
Although the adversarial perturbation generator and the structural component mixup operations are crucial for guiding model training, they are not used during inference. During inference, the fine-tuned foundation model $f$ generates predictions directly through its learned weights, ensuring that there is no additional computational overhead.

\section{Experiments}

\subsection{Experimental Details}

\textbf{Datasets And Metrics.} Our evaluation is conducted on TSFM-Bench \citep{TSFM-2025}, which comprises ten representative forecasting benchmarks, including four ETT subsets, Exchange, ZafNoo, Weather, Electricity (ECL), Solar, and AQShunyi, covering a wide range of real-world scenarios. We quantify predictive performance using Mean Squared Error (MSE) and Mean Absolute Error (MAE). Extensive details of the benchmark datasets are provided in the Appendix~\ref{sec:datasets}.

\textbf{Pre-trained Models And Baselines.}
To evaluate the effectiveness and versatility of our framework, we choose four representative TSFMs as the backbone models: Moirai~\citep{Moirai-2024}, UniTS~\citep{Units-2024}, Timer~\citep{Timer-2024}, and Sundial~\citep{Sundial-2025}. We compare our proposed method with several widely-used fine-tuning and adaptation baselines, including Linear~\citep{linear-2022}, LoRA~\citep{lora-2022}, AdaLoRA~\citep{adalora-2023}, and MSFT~\citep{MSFT-2025}.

\textbf{Implementation Details.}
We follow the experimental protocol of TSFM-Bench~\citep{TSFM-2025}, using prediction horizons $\{96,192,336,720\}$ and selecting the optimal lookback length from $\{96,336,512\}$ based on validation performance.
All methods are trained with Adam~\citep{ADAM-2015} for at most 20 epochs, using an initial learning rate of $1\times10^{-4}$, an annealing schedule, and early stopping with a patience of 3.
For fair comparison, we follow their official implementations and tune the hyperparameters on the validation set to obtain the best configuration.
All reported results are averaged over five independent runs with different random seeds.

\subsection{Experimental Results and Analysis}

\textbf{Main Results.}
\begin{table}[t]
  \caption{Comparative results for long-term multivariate forecasting. Metrics represent the average MSE across horizons $\{96, 192, 336, 720\}$. \textbf{Bold} indicates the best fine-tuning performance for each TSFM, and the overall best across all backbones is in {\color[HTML]{C00000} \textbf{red}}. Our method is highlighted in green.
  See Appendix~\ref{sec:full_mse_timer} for full results.}
  \setlength{\tabcolsep}{3pt}
  \label{tab:main-result}
  \centering
  \renewcommand{\arraystretch}{1.3}   
  \footnotesize
    \resizebox{0.98\linewidth}{!}{
    \begin{tabular}{l|*{12}{c}} 
    \toprule
        \textbf{Method} & \textbf{ETTh1} & \textbf{ETTh2} & \textbf{ETTm1} & \textbf{ETTm2} & \textbf{Exchange} & \textbf{ZafNoo} & \textbf{Weather} & \textbf{Solar} & \textbf{ECL} & \textbf{AQShunyi}& \textbf{Avg} \\
        
        \hline
        \addlinespace[1.5pt]
        PatchTST & 0.469 & 0.387 & 0.387 & 0.269 & 0.450 & 0.512 & 0.265 & 0.207 & 0.216 & 0.808 & 0.397 \\
        iTransformer & 0.439 & 0.370 & 0.361 & 0.269 & 0.360 & 0.522 & 0.258 & 0.207 & 0.182 & 0.706 & 0.367 \\
        TimeMixer & 0.436 & 0.364 & 0.356 & 0.406 & 0.383 & 0.518 & 0.226 & 0.193 & 0.185 & 0.715 & 0.378 \\   
        TimesNet
        & 0.468 & 0.390 & 0.407 & 0.292 & 0.406 & 0.536
        & 0.255 & 0.211 & 0.190 & 0.727 & 0.388 \\
        FedFormer
        & 0.433 & 0.406 & 0.567 & 0.335 & 0.501 & 0.575
        & 0.312 & 0.641 & 0.218 & 0.763 & 0.475 \\

        \hline
        \addlinespace[1.5pt]
        
        \textbf{Timer} & 0.465 & 0.394 & 0.431 & 0.289 & 0.370 & 0.645 & 0.271 & 0.225 & 0.168 & 0.788 & 0.405  \\
        + \textit{LoRA} & 0.452 & 0.382 & 0.792 & 0.318 & 0.367 & 0.612 & 0.537 & 0.345 & 0.421 & 0.989 & 0.522  \\ 
        + \textit{AdaLoRA} & 0.457 & 0.380 & 0.764 & 0.313 & 0.369 & 0.585 & 0.391 & 0.305 & 0.265 & 0.774 & 0.460 \\
        + \textit{Linear} & 0.456 & 0.377 & 0.808 & 0.325 & 0.358 & 0.649 & 0.380 & 0.280 & 0.222 & 0.767 & 0.462  \\
        + \textit{MSFT} & 0.421 & 0.391 & 0.364 & 0.271 & 0.446 & \textbf{0.515} & 0.244 & 0.209 & 0.181 & 0.750 & 0.379 \\

        \rowcolor{oursgreen}
        + \textit{\textbf{Ours}} &  \color[HTML]{C00000} \textbf{0.390} & \textbf{0.376} &  \color[HTML]{C00000} \textbf{0.342} & \color[HTML]{C00000} \textbf{0.262} & \color[HTML]{C00000} \textbf{0.351} & \textbf{0.515} & \color[HTML]{C00000} \textbf{0.225} & \textbf{0.207} & \color[HTML]{C00000} \textbf{0.164} & \textbf{0.733}  & \color[HTML]{C00000} \textbf{0.357} \\
        \hline
        \addlinespace[1.5pt]
        
        \textbf{UniTS}  & 0.450 & 0.408 & 0.422 & 0.313 & 0.396 & 0.523 & 0.245 & 0.188 & 0.166 & 0.802 & 0.391  \\
        + \textit{LoRA} & 0.458 & 0.403 & 0.432 & 0.327 & 0.389 & 0.631 & 0.250 & 1.866 & 0.188 & 0.897 & 0.584  \\
        + \textit{AdaLoRA} & 0.436 & 0.384 & 0.392 & 0.309 & 0.366 & 0.566 & 0.256 & 0.313 & 0.180 & 0.820 & 0.402  \\
        + \textit{Linear} & 0.438 & 0.399 & 0.392 & 0.330 & 0.367 & 0.614 & 0.273 & 0.215 & 0.200 & 0.820 & 0.405  \\
        + \textit{MSFT}& 0.434 & 0.380 & 0.390 & 0.286 & 0.360 & 0.557 & 0.241 & 0.237 & 0.184 & 0.807  & 0.388 \\

        \rowcolor{oursgreen}
        + \textit{\textbf{Ours}} & \textbf{0.424} &  \color[HTML]{C00000} \textbf{0.361} & \textbf{0.380} & \textbf{0.285} & \textbf{0.355} & \textbf{0.508} & \textbf{0.240} & \color[HTML]{C00000} \textbf{0.177} &  \color[HTML]{C00000} \textbf{0.164} & \textbf{0.785} & \textbf{0.368} \\
        \hline

        \addlinespace[1.5pt]
        \textbf{MoiRai}  & 0.522 & 0.382 & 0.424 & 0.340 & 0.452 & 0.598 & 0.246 & 0.205 & 0.259 & 0.725 & 0.415  \\
        + \textit{LoRA} & 0.574 & 0.431 & 0.443 & 0.323 & 0.460 & 0.665 & 0.350 & 1.868 & 0.173 & 0.948 & 0.624 \\
        + \textit{AdaLoRA} & 0.540 & 0.372 & 0.432 & 0.292 & 0.567 & 0.576 & 0.243 & 0.206 & 0.173 &  0.653& 0.405   \\
        + \textit{Linear} & 0.520 & 0.385 & 0.445 & 0.293 & 0.485 & 0.590 & 0.251 & 0.284 & 0.182 &  \color[HTML]{C00000} \textbf{0.599} & 0.404 \\
        + \textit{MSFT} & \textbf{0.475} & 0.372 & 0.417 & 0.295 & 0.390 & 0.558 & 0.241 & 0.207 & 0.173 & 0.686 & 0.381 \\

        \rowcolor{oursgreen}
        + \textit{\textbf{Ours}} & 0.489 & \textbf{0.370} & \textbf{0.407} & \textbf{0.284} & \textbf{0.385} & \textbf{0.524} & \textbf{0.238} & \textbf{0.196} & \textbf{0.171} & 0.713 & \textbf{0.378} \\
        \hline

        \addlinespace[1.5pt]
        \textbf{Sundial} & 0.455 & 0.592 & 0.402 & 0.477 & 0.472 & 0.515 & 0.259 & 0.211 & 0.188 & 0.821 & 0.439  \\
        + \textit{LoRA} & 0.468 & \textbf{0.517} & 0.435 & 0.459 & 0.473 & 0.647 & 0.308 & 1.921 & 0.288 & 0.985 & 0.650 \\
        + \textit{AdaLoRA} & 0.512 & 0.776 & 0.642 & 0.753 & 0.510 & 0.594 & 0.456 & 0.308 & 0.195 & \textbf{0.712}   & 0.546  \\
        + \textit{Linear} & 0.560 & 0.870 & 0.627 & 0.521 & 0.504 & 0.565 & 0.278 & 0.334 & 0.199 & 0.715 & 0.517 \\
        + \textit{MSFT} & 0.598 & 1.066 & 0.425 & 0.424 & 0.557 & 0.629 & 0.259 & 0.259 & 0.191 & 0.718  & 0.513 \\

        \rowcolor{oursgreen}
        + \textit{\textbf{Ours}} & \textbf{0.450} & 0.550 & \textbf{0.396} & \textbf{0.387} & \textbf{0.363} & \color[HTML]{C00000} \textbf{0.486} & \textbf{0.253} & \textbf{0.204} & \textbf{0.187} & \textbf{0.712} & \textbf{0.399} \\
    \bottomrule
    \end{tabular}
    }
\end{table}
Table~\ref{tab:main-result} compares our method with various baselines across 10 datasets, 4 foundation models, and 4 prediction horizons of $\{96,192,336,720\}$.
SIFT achieves the best fine-tuning performance in the vast majority of evaluated backbone-dataset settings, outperforming full fine-tuning and existing adaptation baselines across a broad range of scenarios.
Notably, it reduces the average MSE of Timer by 11.9\% compared with full fine-tuning, while lower average MSE is also achieved on UniTS, Moirai, and Sundial. 
These results demonstrate that SIFT generalizes well across diverse TSFM architectures.
Across datasets, SIFT also demonstrates consistent improvements over a broad range of forecasting benchmarks. For instance, compared with full fine-tuning, SIFT achieves 17.0\% lower MSE on Weather, 9.3\% on ETTm2, and 8.0\% on Solar with Timer. 
On Exchange, it reduces MSE by 10.4\% and 14.8\% for UniTS and Moirai, respectively.

\textbf{Ablation Study.}
We evaluate the contributions of SIAA and CSFE by progressively replacing their key components with simpler alternatives.
As shown in Table~\ref{tab:ablation}, for SIAA, we compare random augmentation, unconstrained adversarial perturbation, and adversarial perturbation with semantic subspace projection.
Random augmentation yields only marginal gains, while unconstrained adversarial perturbation is unstable across datasets.
Adding the semantic projection consistently improves performance, demonstrating SIAA's effectiveness in improving semantic-invariant robustness.
For CSFE, we compare naive mixup with component-aware mixup and reconstruction.
The latter achieves better performance, showing the benefit of explicitly preserving temporal structure.
Finally, combining SIAA and CSFE yields the best overall results, demonstrating their complementary effects.

\begin{table}[t]
\centering
\caption{Ablation study of the proposed Semantic-Invariant Adversarial Augmentation (SIAA) and Component-based Structural Fidelity Enhancement (CSFE).}
\label{tab:ablation}
\setlength{\tabcolsep}{4pt}
\begin{tabular}{lcc|cccc}
\toprule
Method & SIAA Variant& CSFE Variant
& ETTh1 & ETTm1 & Exchange & ZafNoo \\
\midrule
Full-FT & -- & -- & 0.450 & 0.422 & 0.396 & 0.523 \\
+ Random Aug. & Random & -- & 0.431 & 0.389 & 0.386 & 0.521 \\
+ Adv. w/o Proj. & Adv. & -- & 0.455 & 0.385 & 0.370 & 0.554 \\
+ SIAA & Adv. + Proj. & -- & 0.427 & 0.382 & 0.360 & 0.517 \\
\midrule
+ Naive Mixup & -- & Mixup & 0.450 & 0.390 & 0.399 & 0.542 \\
+ CSFE & -- & Mixup + Decomp. & 0.428 & 0.386 & 0.390 & 0.514 \\
\midrule
\textbf{Ours Full}
& \textbf{Adv. + Proj.}
& \textbf{Mixup + Decomp.}
& \textbf{0.424} & \textbf{0.380} & \textbf{0.355} & \textbf{0.508} \\
\bottomrule
\end{tabular}
\end{table}

\begin{figure}[t]
  \centering
  \includegraphics[width=0.98\textwidth]{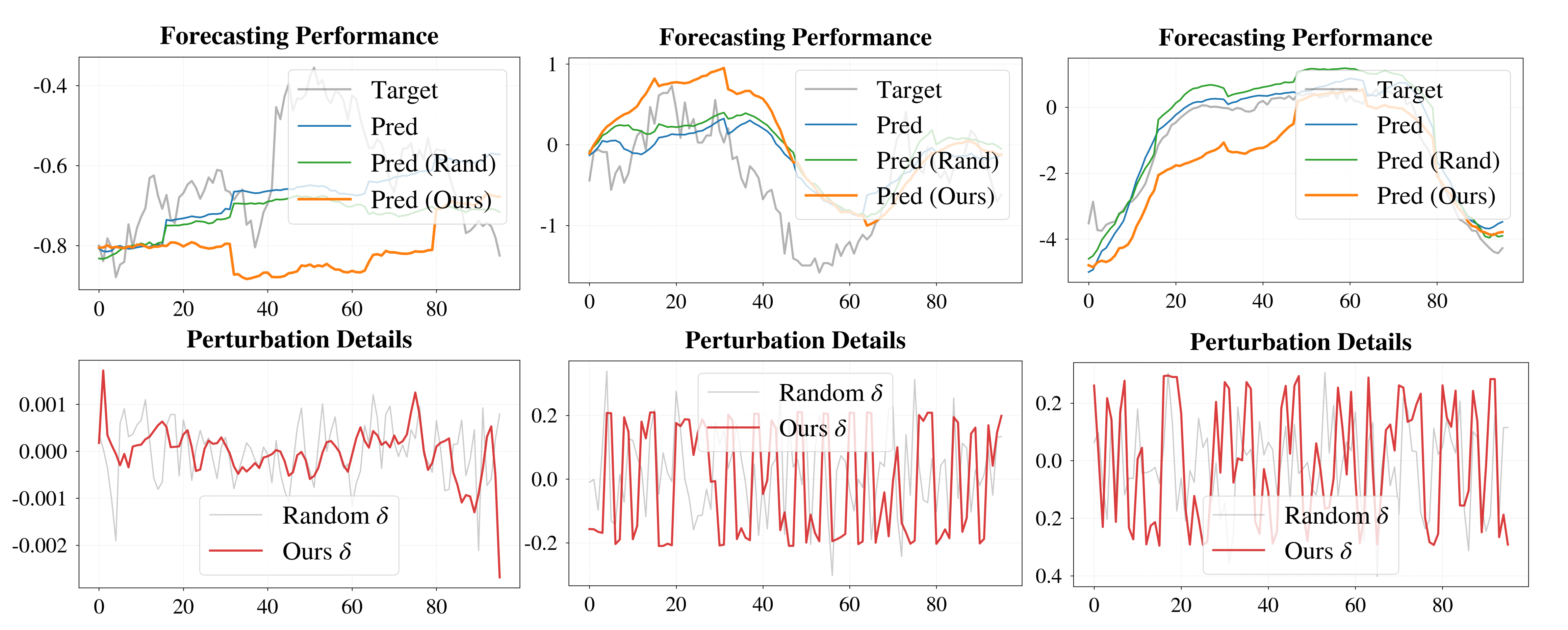} 
  \caption{Effectiveness analysis of perturbation generator.}
  \label{fig:adv_effectiveness_case}
\end{figure}

\textbf{Effectiveness of the adversarial perturbation generator.}
We conduct a comparison on a fully fine-tuned UniTS checkpoint to evaluate the perturbations produced by our adversarial generator $PG$ against stochastic noise.
As illustrated in Figure~\ref{fig:adv_effectiveness_case}, with both perturbations constrained to the same magnitude, the learned perturbations induce substantially larger deviations from the original forecasts than stochastic noise.
This suggests that $P_G$ identifies model-sensitive temporal variations rather than introducing arbitrary disturbances.

\begin{figure}[t] 
  \centering
  \begin{subfigure}[t]{0.24\textwidth}
      \centering
      \includegraphics[width=\textwidth]{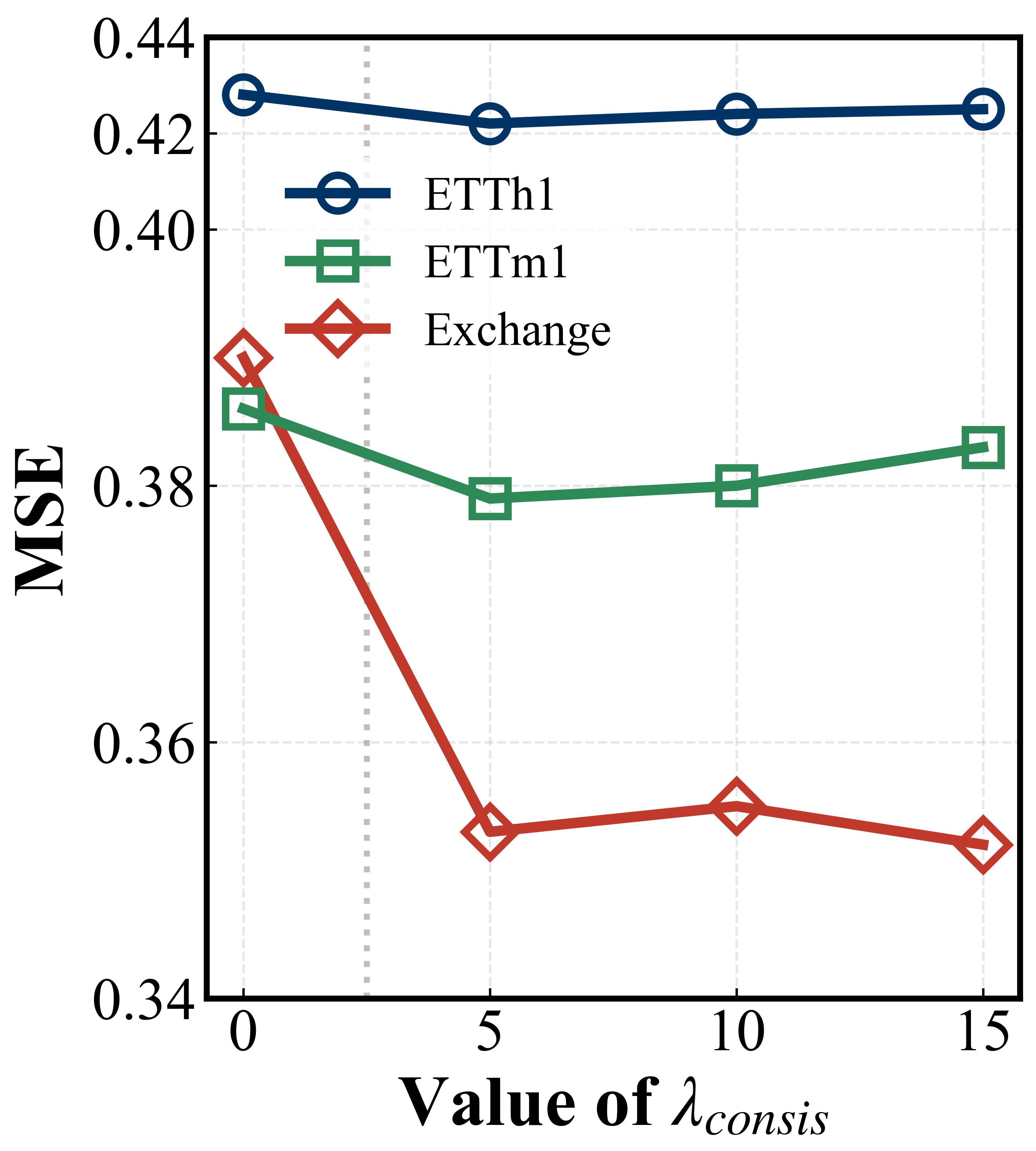}
      \caption{Sensitivity of $\lambda_{\text{consis}}$}
      \label{fig:sens_a}
  \end{subfigure}
  \hfill
  \begin{subfigure}[t]{0.24\textwidth}
      \centering
      \includegraphics[width=\textwidth]{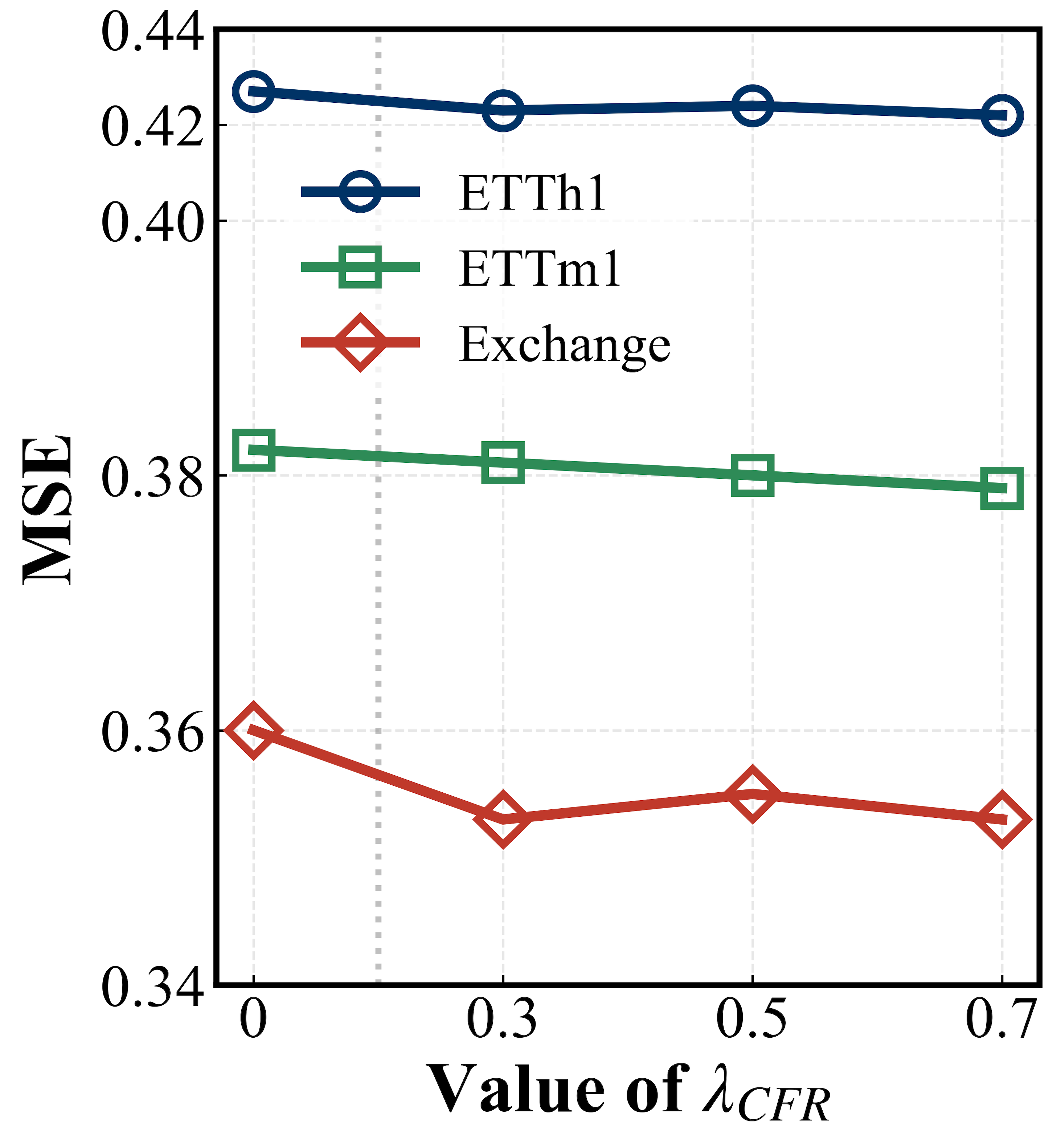}
      \caption{Sensitivity of $\lambda_{\text{CFR}}$ }
      \label{fig:sens_b}
  \end{subfigure}
  \hfill
  \begin{subfigure}[t]{0.24\textwidth}
      \centering
      \includegraphics[width=\textwidth]{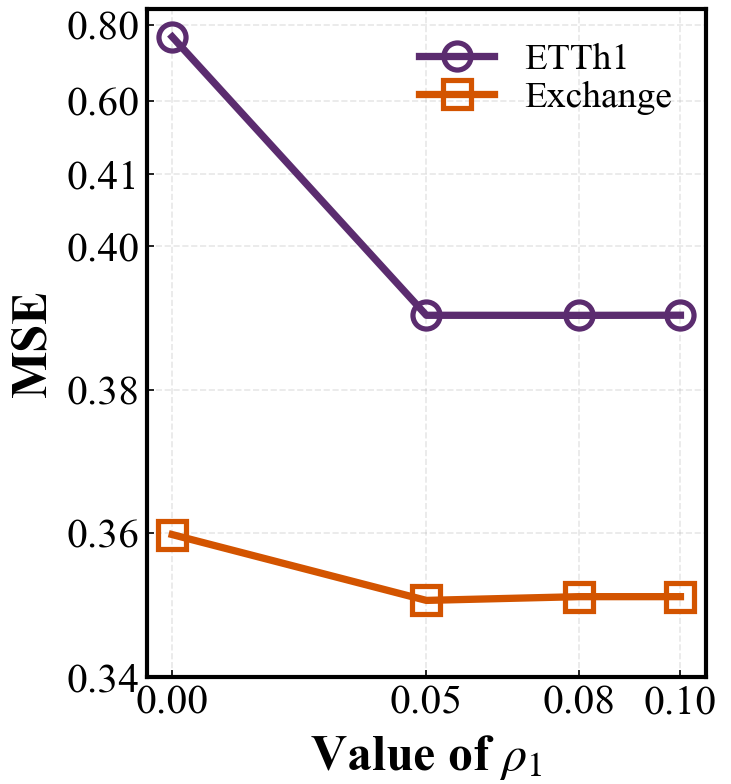}
      \caption{Sensitivity of $\rho_1$}
      \label{fig:rank1}
  \end{subfigure}
  \hfill
  \begin{subfigure}[t]{0.24\textwidth}
      \centering
      \includegraphics[width=\textwidth]{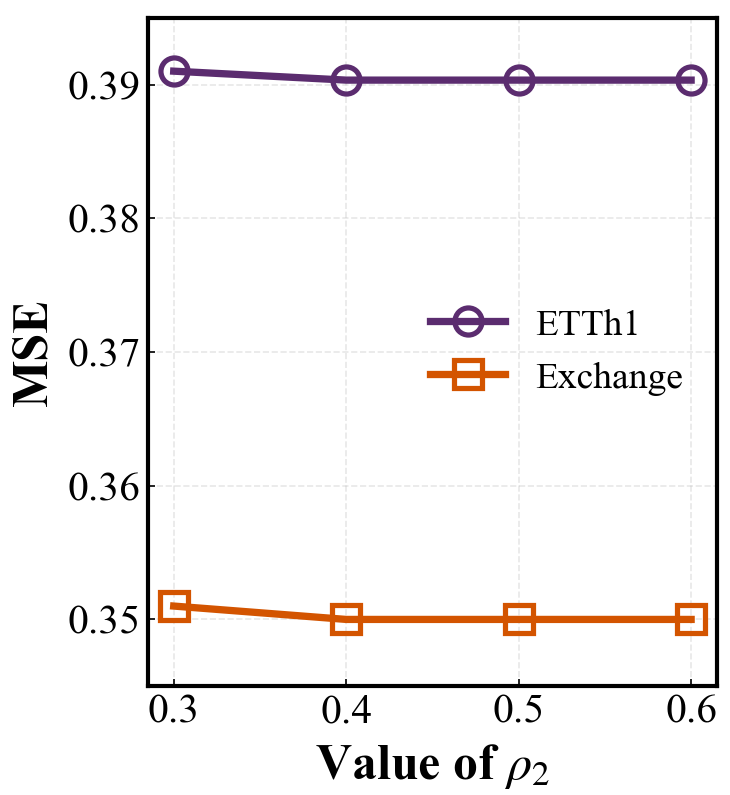}
      \caption{Sensitivity of $\rho_2$}
      \label{fig:sens_d}
  \end{subfigure}

  \caption{Parameter sensitivity analysis across different datasets.}
  \label{fig:overall_sensitivity}
\end{figure}

\textbf{Sensitivity Analysis.}
To analyze the influence of the loss coefficients, we conduct a sensitivity analysis by varying $\lambda_{\text{consis}}$ and $\lambda_{\text{CFR}}$ independently. 
We evaluate $\lambda_{\text{consis}} \in \{0, 5, 10, 15\}$ and $\lambda_{\text{CFR}} \in \{0, 0.3, 0.5, 0.7\}$. 
As shown in Figure~\ref{fig:sens_a} and~\ref{fig:sens_b}, the sensitivity analysis reveals that the forecasting results remain stable across a wide range of non-zero values of $\lambda_{\text{consis}}$ and $\lambda_{\text{CFR}}$. 
The model achieves its best performance at $\lambda_{\text{consis}}=5$ and $\lambda_{\text{CFR}}=0.7$. 
These results indicate that both losses contribute meaningfully to performance, while the model is not sensitive to the exact choice of loss weights within a reasonable range.

\textbf{Impact of Rank Selection Ratios.}
We evaluate the sensitivity of rank ratios by partitioning the singular spectrum into a core semantic space ($< r_1$), a subordinate semantic perturbation zone ($[r_1, r_2]$), and a residual noise space ($> r_2$). 
Following the definition in ~\ref{sec:SSD}, the partition boundaries are governed by the ratios $\rho_1$ and $\rho_2$ such that $r_1 = \lfloor d \cdot \rho_1 \rfloor$ and $r_2= \lfloor d \cdot \rho_2 \rfloor$, where $d = \min(W, K)$ denotes the maximum possible rank.  We vary $\rho_1 \in \{0, 0.05, 0.08, 0.1\}$ and $\rho_2 \in \{0.3, 0.4, 0.5, 0.6\}$. As shown in Figure~\ref{fig:rank1} and ~\ref{fig:sens_d}, the model suffers a huge performance drop at $\rho_1 = 0$, underscoring the necessity of preserving the core semantic backbone. Once $\rho_1 > 0$, the performance remains broadly stable within a wide range of $\rho_1$ and $\rho_2$ configurations, demonstrating that our method is robust to the selection of rank thresholds.

\section{Conclusion}
We introduce SIFT, a novel fine-tuning method designed to address the challenge of overfitting and the mean-prediction trap in TSFMs, which are primarily caused by the low-dimensional nature and diverse structural patterns of time series data. Our method employs semantic-invariant adversarial augmentation to enhance the model’s robustness against semantic-preserving perturbations, mitigating overfitting; simultaneously, it adopts a component-based structural fidelity enhancement to improve the model's ability to retain fine-grained dynamics, alleviating the mean-prediction trap. Our experiments demonstrate that SIFT can significantly boost the fine-tuning performance of TSFMs.

\subsection*{AI use statement}


Generative AI tools were used only as writing assistance during the preparation of this manuscript, mainly for language polishing, grammatical refinement, and improving readability.
The research problem, technical motivation, methodological design, experimental setup, and scientific conclusions were developed by us independently.
All AI-assisted text was manually inspected, revised, and verified to ensure technical accuracy and consistency with the authors' intended claims.
We are fully responsible for the content of the final manuscript, including all statements and materials that were prepared with the assistance of generative AI.



\subsection*{Ethics statement}
This study uses only publicly available time-series benchmark datasets and does not involve human participants, private records, or personally identifiable information.
The proposed method is developed for general-purpose time-series forecasting and does not rely on sensitive attributes or user-specific data.
We are not aware of any substantial ethical risks arising directly from the methodology or experimental setting considered in this work.

\subsection*{Reproducibility statement}
We describe the experimental protocol, model configurations, training settings, and evaluation procedures in the main paper and supplementary material.
The datasets evaluated in this work are publicly accessible, and the experiments follow the benchmark settings specified in the paper.
To facilitate independent reproduction, we include the source code and instructions for reproducing the experiments in the supplementary material.

\bibliography{iclr2027_conference}
\bibliographystyle{iclr2027_conference}

\newpage
\appendix

\section{APPENDIX}

\subsection{Datasets}
To conduct a rigorous and unbiased assessment of our models, we evaluate their performance across ten well-established forecasting benchmarks. These datasets encompass a diverse range of domains: (1) the ETT suite~\citep{ETT-2021}, which captures seven variates from two electric transformers between 2016 and 2018 across four subsets of varying granularity (hourly ETTh1/ETTh2 and 15-minute ETTm1/ETTm2); (2) the Electricity dataset~\citep{electricity-2015}, providing hourly consumption logs for 321 customers from 2016 to 2019; (3) the Exchange dataset~\citep{solar-2018}, which tracks daily exchange rates for eight countries from 1990 to 2016; (4) the Solar dataset~\citep{solar-2018}, featuring 10-minute interval power generation from 137 PV plants in 2006; (5) the Weather dataset~\citep{autoformer-2021}, documenting 21 meteorological indicators for Germany in 2020 at 10-minute intervals; (6) the AQShunyi dataset\citep{AQShunyi-2017} which offers comprehensive multi-year air quality measurements; and (7) the ZafNoo dataset\citep{zafnoo-2020}, which provides sap flow data along with associated environmental variables. Comprehensive details regarding these benchmarks are provided in the table~\ref{tab:statistics_of_datasets}.

\label{sec:datasets}
\begin{table}[h]
  \centering
  \renewcommand{\arraystretch}{1.2} 
  \caption{Statistics of the benchmark datasets used in our experiments.}
  \label{tab:statistics_of_datasets}
  \begin{tabular}{l l c r r c}
    \toprule
    Dataset & Domain & Frequency & Samples & Dim & Train/Val/Test \\
    \midrule
    ETTh1 & Electricity & 1h & 14,400 & 7 & 6:2:2 \\
    ETTh2 & Electricity & 1h & 14,400 & 7 & 6:2:2 \\
    ETTm1 & Electricity & 15m & 57,600 & 7 & 6:2:2 \\
    ETTm2 & Electricity & 15m & 57,600 & 7 & 6:2:2 \\
    Electricity & Electricity & 1h & 26,304 & 321 & 7:1:2 \\
    Solar & Energy & 10m & 52,560 & 137 & 6:2:2 \\
    Weather & Environment & 10m & 52,696 & 21 & 7:1:2 \\
    AQShunyi & Environment & 1h & 35,064 & 11 & 6:2:2 \\
    ZafNoo & Nature & 30m & 19,225 & 11 & 7:1:2 \\
    Exchange & Economic & 1d & 7,588 & 8 & 7:1:2 \\
    \bottomrule
  \end{tabular}
\end{table}

\subsection{TSFMs}

\textbf{Timer}~\citep{Timer-2024}: Timer is a GPT-style autoregressive model for time series analysis that treats time series forecasting as a next-token prediction task. By training on large-scale time series datasets, it demonstrates strong sequence modeling capabilities, effectively supporting various downstream tasks such as forecasting, imputation, and anomaly detection across diverse domains.

\textbf{UniTS}~\citep{Units-2024}: UniTS is a unified time series model that supports general task specifications, accommodating classification, forecasting, imputation, and anomaly detection. This is achieved through a novel unified backbone that integrates both sequential and variate attention with dynamic linear operators, all trained as a single unified model.

\textbf{Moirai}~\citep{Moirai-2024}: Moirai is a universal time series forecasting model based on a patch-based~\citep{patch-2022} Transformer architecture. It addresses three unique challenges of time series foundation models: cross-frequency learning, accommodating an arbitrary number of variates, and handling varying distributional properties in large-scale data. 

\textbf{Sundial}~\citep{Sundial-2025}: Sundial is a family of time series foundation models that introduces TimeFlow Loss, a flow-matching based training objective that enables native pre-training of Transformers on continuous-valued time series without discrete tokenization. Unlike prior methods that rely on parametric densities, Sundial learns arbitrary distributions flexibly and can generate multiple probable predictions. Sundial achieves state-of-the-art zero-shot performance on both point and probabilistic forecasting benchmarks, with efficient inference speed.

\subsection{Fine-tuning baselines}

\textbf{Full Fine-tune and Linear Probe.} 
Full fine-tuning updates all parameters of the pretrained model. We use a shared learning rate for both full fine-tuning and linear probing. Linear probing~\citep{linear-2022} only updates the output head while keeping the backbone frozen.

\textbf{LoRA and AdaLoRA.} 
LoRA~\citep{lora-2022} introduces trainable low-rank decomposition matrices into attention layers, enabling parameter-efficient fine-tuning by injecting updates into a low-rank subspace. AdaLoRA~\citep{adalora-2023} further improves this process by dynamically allocating the rank during training based on parameter importance.
In our implementation, we apply LoRA and AdaLoRA to the query, key, and value projection layers. For LoRA and AdaLoRA, we follow the official configurations as the starting point and tune the key hyperparameters on the validation set.

\textbf{MSFT.} 
For MSFT~\citep{MSFT-2025}, we follow the official implementation and tune its key hyperparameters on the validation set. The selected configuration uses rank \(r=16\) and scaling number \(scale\_num=3\).

\subsection{Visualizations of Semantic Spectrum Decomposition}
\label{sec:SSA}

\begin{figure}[t]
    \centering
    \includegraphics[width=0.98\textwidth]{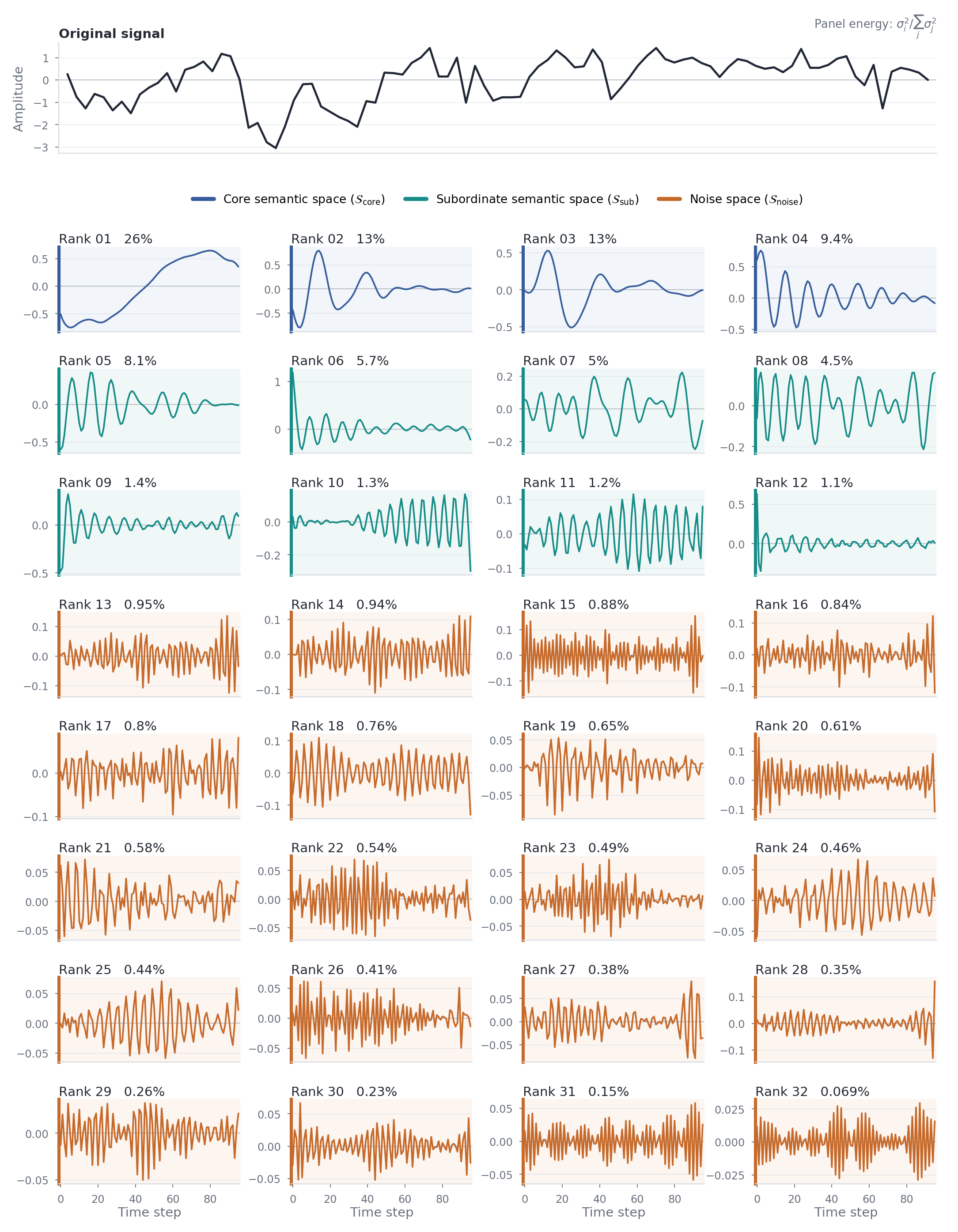}
    \caption{
    Rank-wise visualization of the proposed spectral decomposition on an ETTh1 sample.
    The top panel shows the original time series, while the remaining panels visualize components reconstructed from individual singular directions together with their relative spectral energy.
    Low-rank components capture dominant large-scale temporal structure, intermediate components preserve structured but non-dominant variations, and high-rank components mainly contain weak high-frequency fluctuations.
    Colors denote the core semantic space ($\mathcal{S}_{\mathrm{core}}$), subordinate semantic space ($\mathcal{S}_{\mathrm{sub}}$), and noise space ($\mathcal{S}_{\mathrm{noise}}$), respectively.
    }
    \label{fig:ssd}
\end{figure}

To provide an intuitive interpretation of the proposed spectral decomposition, we visualize the rank-wise reconstructed components of an ETTh1 sample in Figure~\ref{fig:ssd}. Each component is reconstructed from an individual singular direction and is annotated with its relative spectral energy. The leading components account for the majority of the signal energy and exhibit smooth, large-scale temporal variations, capturing the dominant trend and long-period seasonal patterns of the original sequence. 

As the rank increases, the reconstructed components gradually transition toward higher-frequency and more localized oscillations. In particular, the intermediate-rank components retain structured and non-negligible temporal variations while contributing substantially less energy than the dominant components, corresponding to the subordinate semantic space used for adversarial perturbation. 

In contrast, the high-rank components have substantially smaller energy contributions and predominantly exhibit irregular, high-frequency fluctuations, which are assigned to the noise space.

Overall, the visualization reveals a progressive transition from dominant trend and periodic structure to structured intermediate variations and finally to weak and noisy high-frequency fluctuations across the singular spectrum, providing an intuitive interpretation of the three spectral subspaces adopted in SIFT. 
In practice, the partition boundaries are selected on the validation set and then fixed for downstream fine-tuning and evaluation.

\subsection{Semantic Interpretation of Spectral Subspaces}
While the rank-wise visualization above reveals distinct temporal characteristics across the spectral spectrum, it remains important to examine whether the decomposed subspaces capture domain-related temporal information rather than merely reflecting differences in frequency or energy.
We therefore measure the affinity between corresponding subspaces across related ETT datasets and external datasets.

\begin{table}[t]
\centering
\caption{
Cross-domain affinity between the decomposed spectral subspaces.
Higher affinity indicates stronger similarity between the corresponding subspaces.
}
\label{tab:subspace_affinity}
\begin{tabular}{lccc}
\toprule
\textbf{Dataset comparison}
& \textbf{Core}
& \textbf{Subordinate}
& \textbf{Noise} \\
\midrule
ETTh1--ETTh2              & 0.993 & 0.850 & 0.940 \\
ETTm1--ETTm2              & 0.999 & 0.978 & 0.994 \\
Avg. of related ETT pairs & 0.995 & 0.915 & 0.967 \\
Avg. of ETT--external pairs & 0.812 & 0.855 & 0.946 \\
\midrule
Related--external gap     & \textbf{0.183} & 0.060 & 0.021 \\
\bottomrule
\end{tabular}
\end{table}

As shown in Table~\ref{tab:subspace_affinity}, the core subspace exhibits very high affinity across related ETT datasets, while its affinity decreases substantially when comparing ETT with external datasets.
This yields a related--external affinity gap of 0.183, markedly larger than those of the subordinate (0.060) and noise (0.021) subspaces.
These results suggest that the dominant spectral directions capture more domain-shared temporal structure, whereas the subordinate subspace retains more dataset-specific but still structured variations.
The noise subspace, in contrast, shows little discrimination between related and external datasets.

Together with the rank-wise reconstruction in Figure~\ref{fig:ssd}, this analysis indicates that the proposed decomposition is not merely an energy-based partition of the spectrum, but separates temporal components with distinct structural and domain-related characteristics.
This provides empirical support for preserving the dominant core structure while applying adversarial perturbations within the subordinate subspace.

\subsection{Forecasting Visualization.}
\begin{figure}[t]
  \centering
  \includegraphics[width=0.98\textwidth]{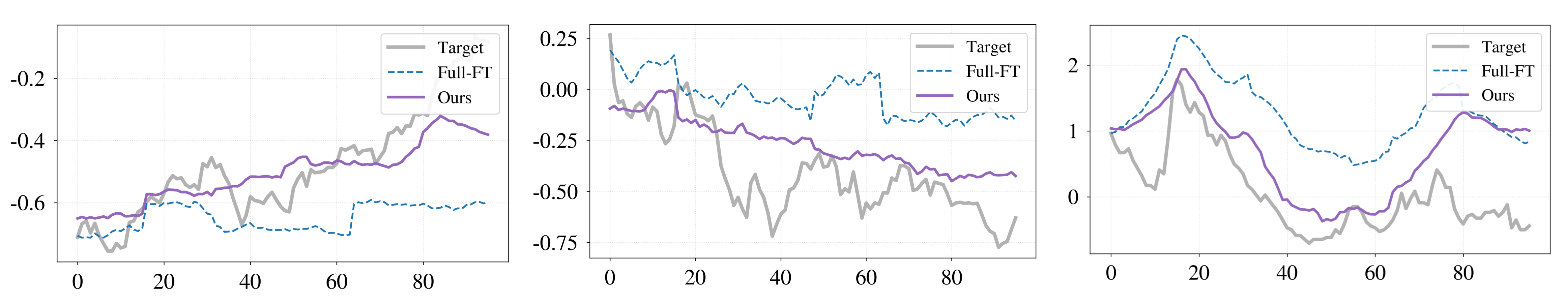} 
  \caption{Forecasting visualization comparing our method and standard fine-tuning. The purple line denotes our method, while the blue line represents Full-FT.}
  \label{fig:case}
\end{figure}

Figure~\ref{fig:case} illustrates the visualization results of representative samples from the ETTm1 and Exchange datasets under both standard fine-tuning and our proposed method. We can observe that our approach aligns more closely with the ground truth. This is primarily because our method enables the model to anchor the primary semantics while maintaining structural fidelity.

\subsection{Limitation of Series Decomposition Assumption}
Although series decomposition techniques are widely adopted in time series modeling, they provide only an approximate separation of trend and seasonal components rather than a fully identifiable or exact factorization. In practice, the decomposition quality may vary across datasets, particularly for highly non-stationary or multi-scale signals where temporal patterns are strongly entangled. As a result, the effectiveness of the proposed structural fidelity enhancement may be influenced by the accuracy of the underlying decomposition.

\FloatBarrier
\subsection{Full Result}
\label{sec:full_mse_timer}
\begin{table}[t] 
  \centering
  \renewcommand{\arraystretch}{1.1} 
  \caption{Full MSE results on Timer and UniTS.}
  \label{tab:full_mse_timer}
  \resizebox{1.0\textwidth}{!}{
  {
    \footnotesize
    \begin{tabular}{clllllllllllll}
    \toprule 
     & \multicolumn{1}{c}{} & \multicolumn{6}{c}{\textbf{Timer}} & \multicolumn{6}{c}{\textbf{UniTS}} \\
     \midrule  
    \multicolumn{1}{l}{} &  & \textit{Full} & \textit{LoRA} & \textit{AdaLoRA} & \textit{Linear} & \textit{MSFT} & \textit{Ours} & \textit{Full} & \textit{LoRA} & \textit{AdaLoRA} & \textit{Linear} & \textit{MSFT} & \textit{Ours} \\
    \midrule  
    \multirow{5}{*}{\textbf{ETTh1}} & 96 & 0.396 & 0.382 & 0.382 & 0.381 & 0.399 & \textbf{0.366} & 0.395 & 0.407 & 0.372 & 0.372 & 0.376 & \textbf{0.371} \\
     & 192 & 0.487 & 0.441 & 0.451 & 0.438 & 0.413 & \textbf{0.384} & 0.446 & 0.443 & 0.421 & 0.419 & 0.428 & \textbf{0.414} \\
     & 336 & 0.483 & 0.487 & 0.479 & 0.486 & 0.426 & \textbf{0.398} & 0.479 & 0.478 & 0.465 & 0.468 & 0.463 & \textbf{0.451} \\
     & 720 & 0.497 & 0.497 & 0.517 & 0.519 & 0.446 & \textbf{0.414} & 0.478 & 0.503 & 0.488 & 0.491 & 0.469 & \textbf{0.459} \\
     & Avg & 0.465 & 0.452 & 0.457 & 0.456 & 0.421 & \textbf{0.390} & 0.450 & 0.458 & 0.436 & 0.438 & 0.434 & \textbf{0.424} \\
     \midrule  
    \multirow{5}{*}{\textbf{ETTh2}} & 96 & 0.305 & 0.295 & 0.293 & \textbf{0.291} & 0.300 & \textbf{0.291} & 0.311 & 0.339 & 0.308 & 0.329 & \textbf{0.302} & 0.303 \\
     & 192 & 0.385 & 0.377 & 0.376 & 0.374 & 0.386 & \textbf{0.373} & 0.470 & 0.416 & 0.387 & 0.414 & 0.375 & \textbf{0.347} \\
     & 336 & 0.437 & 0.418 & 0.423 & \textbf{0.415} & 0.436 & 0.416 & 0.429 & 0.429 & 0.416 & 0.414 & 0.416 & \textbf{0.387} \\
     & 720 & 0.448 & 0.438 & 0.429 & 0.427 & 0.441 & \textbf{0.424} & 0.424 & 0.429 & 0.426 & 0.438 & 0.425 & \textbf{0.406} \\
     & Avg & 0.394 & 0.382 & 0.380 & 0.377 & 0.391 & \textbf{0.376} & 0.408 & 0.403 & 0.384 & 0.399 & 0.380 & \textbf{0.361} \\
     \midrule  
    \multirow{5}{*}{\textbf{ETTm1}} & 96 & 0.335 & 0.327 & 0.327 & 0.327 & 0.308 & \textbf{0.289} & 0.332 & 0.415 & 0.326 & 0.324 & 0.336 & \textbf{0.315} \\
     & 192 & 0.424 & 0.371 & 0.378 & 0.372 & 0.343 & \textbf{0.317} & 0.383 & 0.413 & 0.368 & 0.371 & 0.366 & \textbf{0.360} \\
     & 336 & 0.450 & 1.215 & 1.079 & 1.215 & 0.375 & \textbf{0.351} & 0.467 & 0.431 & 0.408 & 0.420 & 0.396 & \textbf{0.395} \\
     & 720 & 0.514 & 1.255 & 1.271 & 1.319 & 0.432 & \textbf{0.410} & 0.507 & 0.467 & 0.466 & 0.453 & 0.461 & \textbf{0.450} \\
     & Avg & 0.431 & 0.792 & 0.764 & 0.808 & 0.364 & \textbf{0.342} & 0.422 & 0.432 & 0.392 & 0.392 & 0.390 & \textbf{0.380} \\
     \midrule  
    \multirow{5}{*}{\textbf{ETTm2}} & 96 & 0.185 & 0.186 & 0.177 & 0.187 & 0.189 & \textbf{0.165} & 0.198 & 0.242 & 0.190 & 0.193 & 0.179 & \textbf{0.174} \\
     & 192 & 0.257 & 0.260 & 0.246 & 0.262 & 0.237 & \textbf{0.227} & 0.252 & 0.282 & 0.258 & 0.280 & 0.248 & \textbf{0.246} \\
     & 336 & 0.313 & 0.353 & 0.354 & 0.359 & 0.287 & \textbf{0.284} & 0.334 & 0.341 & 0.335 & 0.352 & 0.313 & \textbf{0.310} \\
     & 720 & 0.402 & 0.474 & 0.476 & 0.492 & \textbf{0.372} & \textbf{0.372} & 0.468 & 0.442 & 0.451 & 0.497 & \textbf{0.405} & 0.410 \\
     & Avg & 0.289 & 0.318 & 0.313 & 0.325 & 0.271 & \textbf{0.262} & 0.313 & 0.327 & 0.309 & 0.330 & 0.286 & \textbf{0.285} \\
     \midrule  
    \multirow{5}{*}{\textbf{Exchange}} & 96 & 0.098 & 0.088 & 0.087 & 0.084 & 0.094 & \textbf{0.082} & 0.106 & 0.143 & 0.090 & 0.093 & 0.090 & \textbf{0.085} \\
     & 192 & 0.181 & 0.180 & 0.183 & 0.175 & 0.256 & \textbf{0.172} & 0.222 & 0.218 & 0.187 & 0.199 & 0.185 & \textbf{0.181} \\
     & 336 & 0.332 & 0.337 & 0.336 & 0.320 & 0.421 & \textbf{0.314} & 0.365 & 0.357 & 0.345 & 0.335 & 0.344 & \textbf{0.326} \\
     & 720 & 0.870 & 0.863 & 0.871 & 0.853 & 1.012 & \textbf{0.834} & 0.891 & 0.840 & 0.842 & 0.839 & \textbf{0.820} & 0.827 \\
     & Avg & 0.370 & 0.367 & 0.369 & 0.358 & 0.446 & \textbf{0.351} & 0.396 & 0.389 & 0.366 & 0.367 & 0.360 & \textbf{0.355} \\
     \midrule  
    \multirow{5}{*}{\textbf{ZafNoo}} & 96 & 0.503 & 0.481 & 0.479 & 0.479 & 0.441 & \textbf{0.431} & 0.448 & 0.550 & 0.460 & 0.583 & 0.465 & \textbf{0.431} \\
     & 192 & 0.615 & 0.569 & \textbf{0.481} & 0.563 & 0.490 & 0.484 & 0.492 & 0.614 & 0.538 & 0.606 & 0.524 & \textbf{0.489} \\
     & 336 & 0.682 & 0.636 & 0.630 & 0.629 & \textbf{0.534} & \textbf{0.534} & 0.545 & 0.642 & 0.585 & 0.585 & 0.565 & \textbf{0.537} \\
     & 720 & 0.781 & 0.762 & 0.750 & 0.925 & \textbf{0.595} & 0.612 & 0.608 & 0.716 & 0.682 & 0.683 & 0.674 & \textbf{0.575} \\
     & Avg & 0.645 & 0.612 & 0.585 & 0.649 & \textbf{0.515} & \textbf{0.515} & 0.523 & 0.631 & 0.566 & 0.614 & 0.557 & \textbf{0.508} \\
     \midrule  
    \multirow{5}{*}{\textbf{Weather}} & 96 & 0.172 & 0.498 & 0.174 & 0.176 & 0.178 & \textbf{0.148} & 0.146 & 0.163 & 0.173 & 0.164 & 0.158 & \textbf{0.145} \\
     & 192 & 0.235 & 0.536 & 0.442 & 0.337 & 0.210 & \textbf{0.193} & \textbf{0.200} & 0.214 & 0.221 & 0.223 & 0.204 & 0.203 \\
     & 336 & 0.296 & 0.564 & 0.449 & 0.447 & 0.259 & \textbf{0.244} & 0.272 & 0.274 & 0.278 & 0.301 & \textbf{0.261} & \textbf{0.261} \\
     & 720 & 0.380 & 0.551 & 0.499 & 0.561 & 0.328 & \textbf{0.316} & 0.362 & 0.349 & 0.353 & 0.403 & \textbf{0.342} & 0.350 \\
     & Avg & 0.271 & 0.537 & 0.391 & 0.380 & 0.244 & \textbf{0.225} & 0.245 & 0.250 & 0.256 & 0.273 & 0.241 & \textbf{0.240} \\
     \midrule  
    \multirow{5}{*}{\textbf{Electricity}} & 96 & \textbf{0.134} & 0.442 & 0.135 & 0.182 & 0.152 & 0.139 & \textbf{0.133} & 0.163 & 0.152 & 0.171 & 0.154 & \textbf{0.133} \\
     & 192 & 0.155 & 0.369 & 0.269 & 0.201 & 0.167 & \textbf{0.148} & \textbf{0.153} & 0.170 & 0.168 & 0.184 & 0.169 & \textbf{0.153} \\
     & 336 & 0.172 & 0.423 & 0.319 & 0.224 & 0.183 & \textbf{0.165} & 0.175 & 0.192 & 0.183 & 0.202 & 0.186 & \textbf{0.174} \\
     & 720 & 0.211 & 0.448 & 0.336 & 0.280 & 0.223 & \textbf{0.206} & 0.204 & 0.228 & 0.218 & 0.242 & 0.227 & \textbf{0.197} \\
     & Avg & 0.168 & 0.421 & 0.265 & 0.222 & 0.181 & \textbf{0.164} & 0.166 & 0.188 & 0.180 & 0.200 & 0.184 & \textbf{0.164} \\
     \midrule  
    \multirow{5}{*}{\textbf{Solar}} & 96 & 0.183 & 0.369 & 0.207 & 0.222 & 0.191 & \textbf{0.182} & 0.162 & 1.812 & 0.248 & 0.197 & 0.214 & \textbf{0.161} \\
     & 192 & 0.209 & 0.337 & 0.323 & 0.268 & 0.202 & \textbf{0.198} & 0.195 & 1.871 & 0.255 & 0.198 & 0.237 & \textbf{0.171} \\
     & 336 & \textbf{0.210} & 0.338 & 0.339 & 0.300 & 0.216 & 0.211 & 0.194 & 1.929 & 0.274 & 0.217 & 0.246 & \textbf{0.181} \\
     & 720 & 0.297 & 0.338 & 0.350 & 0.329 & \textbf{0.226} & 0.236 & 0.201 & 1.853 & 0.477 & 0.248 & 0.252 & \textbf{0.196} \\
     & Avg & 0.225 & 0.345 & 0.305 & 0.280 & 0.209 & \textbf{0.207} & 0.188 & 1.866 & 0.313 & 0.215 & 0.237 & \textbf{0.177} \\
     \midrule  
    \multirow{5}{*}{\textbf{AQshunyi}} & 96 & 0.699 & 0.846 & 0.846 & 0.895 & 0.814 & \textbf{0.662} & 0.739 & 0.890 & 0.755 & 0.762 & 0.745 & \textbf{0.722} \\
     & 192 & 0.787 & 1.040 & 0.718 & \textbf{0.687} & 0.693 & 0.765 & 0.784 & 0.941 & 0.816 & 0.817 & 0.805 & \textbf{0.776} \\
     & 336 & 0.809 & 1.034 & 0.738 & \textbf{0.712} & 0.717 & 0.752 & 0.829 & 0.862 & 0.831 & 0.830 & 0.838 & \textbf{0.801} \\
     & 720 & 0.855 & 1.034 & 0.794 & 0.774 & 0.777 & \textbf{0.753} & 0.857 & 0.896 & 0.876 & 0.871 & \textbf{0.842} & 0.843 \\
     & Avg & 0.788 & 0.989 & 0.774 & 0.767 & 0.750 & \textbf{0.733} & 0.802 & 0.897 & 0.820 & 0.820 & 0.807 & \textbf{0.785} \\
     \bottomrule
     
    \end{tabular}
    }
    }
\end{table}

\begin{table}[t] 
  \centering
  \renewcommand{\arraystretch}{1.1} 
  \caption{Full MSE results on Moirai and Sundial.}
  \label{tab:full_mse_timer}
  \resizebox{1.0\textwidth}{!}{
  {
    \footnotesize
    \begin{tabular}{clllllllllllll}
    \toprule 
     & \multicolumn{1}{c}{} & \multicolumn{6}{c}{\textbf{Moirai}} & \multicolumn{6}{c}{\textbf{Sundial}} \\
    \midrule  
    \multicolumn{1}{l}{} &  & \textit{Full} & \textit{LoRA} & \textit{AdaLoRA} & \textit{Linear} & \textit{MSFT} & \textit{Ours} & \textit{Full} & \textit{LoRA} & \textit{AdaLoRA} & \textit{Linear} & \textit{MSFT} & \textit{Ours} \\
    \midrule  
    \multirow{5}{*}{\textbf{ETTh1}} & 96 & 0.495 & 0.575 & \textbf{0.379} & 0.422 & 0.447 & \textbf{0.379} & 0.377 & 0.376 & 0.374 & 0.470 & 0.508 & \textbf{0.370} \\
     & 192 & 0.540 & 0.519 & 0.587 & 0.531 & 0.497 & \textbf{0.418} & 0.422 & 0.493 & 0.443 & 0.537 & 0.571 & \textbf{0.419} \\
     & 336 & 0.491 & 0.500 & 0.463 & 0.543 & \textbf{0.459} & 0.544 & 0.477 & 0.492 & 0.495 & 0.588 & 0.630 & \textbf{0.469} \\
     & 720 & 0.561 & 0.703 & 0.732 & 0.586 & \textbf{0.497} & 0.614 & 0.543 &  \textbf{0.512} & 0.736 & 0.643 & 0.685 & 0.542 \\
     & Avg & 0.522 & 0.574 & 0.540 & 0.520 & \textbf{0.475} & 0.489 & 0.455 & 0.468 & 0.512 & 0.560 & 0.598 & \textbf{0.450} \\
     \midrule  
    \multirow{5}{*}{\textbf{ETTh2}} & 96 & 0.319 & 0.435 & 0.314 & 0.308 & 0.290 & \textbf{0.285} & 0.393 & 0.412 & 0.363 & 0.526 & 0.633 & \textbf{0.343} \\
     & 192 & 0.385 & 0.420 & 0.376 & 0.388 & 0.348 & \textbf{0.345} & 0.548 & \textbf{0.459} & 0.503 & 0.746 & 0.896 & 0.474 \\
     & 336 & 0.391 & 0.395 & 0.408 & 0.379 & 0.381 & \textbf{0.361} & 0.519 & \textbf{0.482} & 0.556 & 0.910 & 1.139 & 0.495 \\
     & 720 & 0.432 & 0.474 & \textbf{0.390
     }& 0.466 & 0.468 & 0.489 & 0.906 & \textbf{0.714} & 1.684 & 1.297 & 1.597 & 0.886 \\
     & Avg & 0.382 & 0.431 & 0.372 & 0.385 & 0.372 & \textbf{0.370} & 0.592 & \textbf{0.517} & 0.776 & 0.870 & 1.066 & 0.550 \\
     \midrule  
    \multirow{5}{*}{\textbf{ETTm1}} & 96 & \textbf{0.312} & 0.410 & 0.358 & 0.373 & 0.368 & 0.336 & 0.334 & 0.407 & 0.806 & 0.575 & 0.363 & \textbf{0.326} \\
     & 192 & 0.422 & 0.417 & 0.393 & 0.412 & \textbf{0.379} & 0.393 & 0.379 & 0.410 & 0.381 & 0.613 & 0.407 & \textbf{0.369} \\
     & 336 & 0.449 & 0.450 & 0.441 & 0.444 & \textbf{0.434} & 0.443 & 0.427 & 0.427 & 0.426 & 0.636 & 0.438 & \textbf{0.415} \\
     & 720 & 0.515 & 0.493 & 0.536 & 0.552 & 0.488 & \textbf{0.456} & \textbf{0.469} & 0.495 & 0.953 & 0.685 & 0.492 & 0.476 \\
     & Avg & 0.424 & 0.443 & 0.432 & 0.445 & 0.417 & \textbf{0.407} & 0.402 & 0.435 & 0.642 & 0.627 & 0.425 & \textbf{0.396} \\
     \midrule  
    \multirow{5}{*}{\textbf{ETTm2}} & 96 & 0.224 & 0.242 & 0.195 & 0.193 & 0.207 & \textbf{0.183} & 0.246 & 0.249 & 0.569 & 0.385 & 0.252 & \textbf{0.210} \\
     & 192 & 0.308 & 0.282 & 0.267 & 0.279 & 0.268 & \textbf{0.239} & 0.354 & \textbf{0.281} & 0.308 & 0.385 & 0.346 & 0.322 \\
     & 336 & 0.369 & 0.341 & 0.310 & \textbf{0.300} & 0.310 & 0.330 & 0.456 & \textbf{0.371} & 0.571 & 0.539 & 0.457 & 0.452 \\
     & 720 & 0.460 & 0.425 & 0.397 & 0.402 & 0.397 & \textbf{0.382} & 0.852 & 0.936 & 1.565 & 0.775 & 0.639 & \textbf{0.563} \\
     & Avg & 0.340 & 0.323 & 0.292 & 0.293 & 0.295 & \textbf{0.284} & 0.477 & 0.459 & 0.753 & 0.521 & 0.424 & \textbf{0.387} \\
     \midrule  
    \multirow{5}{*}{\textbf{Exchange}} & 96 & 0.096 & 0.162 & 0.119 & 0.121 & 0.082 & \textbf{0.081} & 0.120 & 0.160 & 0.221 & 0.155 & 0.204 & \textbf{0.109} \\
     & 192 & 0.200 & 0.229 & 0.249 & 0.292 & 0.182 & \textbf{0.177} & 0.267 & 0.214 & 0.255 & 0.300 & 0.373 & \textbf{0.190} \\
     & 336 & 0.381 & 0.421 & 0.475 & 0.490 & \textbf{0.336} & 0.433 & 0.511 & 0.457 & 0.485 & 0.507 & 0.605 & \textbf{0.365} \\
     & 720 & 1.130 & 1.025 & 1.427 & 1.036 & 0.959 & \textbf{0.849} & 0.988 & 1.061 & 1.076 & 1.054 & 1.046 & \textbf{0.788} \\
     & Avg & 0.452 & 0.460 & 0.567 & 0.485 & 0.390 & \textbf{0.385} & 0.472 & 0.473 & 0.510 & 0.504 & 0.557 & \textbf{0.363} \\
     \midrule  
    \multirow{5}{*}{\textbf{ZafNoo}} & 96 & 0.541 & 0.717 & 0.539 & 0.534 & 0.612 & \textbf{0.424} & 0.478 & 0.761 & 0.570 & 0.484 & 0.539 & \textbf{0.419} \\
     & 192 & 0.604 & 0.646 & 0.582 & 0.542 & \textbf{0.486} & 0.499 & 0.516 & 0.648 & 0.520 & 0.548 & 0.612 & \textbf{0.479} \\
     & 336 & 0.573 & 0.665 & 0.612 & 0.625 & \textbf{0.567} & 0.588 & 0.522 & 0.599 & 0.529 & 0.587 & 0.656 & \textbf{0.505} \\
     & 720 & 0.675 & 0.632 & 0.571 & 0.659 & \textbf{0.565} & 0.586 & 0.545 & 0.581 & 0.758 & 0.639 & 0.710 & \textbf{0.542} \\
     & Avg & 0.598 & 0.665 & 0.576 & 0.590 & 0.558 & \textbf{0.524} & 0.515 & 0.647 & 0.594 & 0.565 & 0.629 & \textbf{0.486} \\
     \midrule  
    \multirow{5}{*}{\textbf{Weather}} & 96 & 0.161 & 0.333 & 0.160 & 0.160 & \textbf{0.148} & 0.149 & 0.171 & 0.276 & 0.329 & 0.198 & 0.181 & \textbf{0.167} \\
     & 192 & \textbf{0.191} & 0.323 & 0.223 & 0.220 & 0.205 & 0.207 & 0.222 & 0.265 & 0.422 & 0.244 & 0.228 & \textbf{0.229} \\
     & 336 & 0.278 & 0.350 & 0.251 & 0.264 & 0.270 & \textbf{0.247} & 0.281 & 0.296 & 0.478 & 0.296 & 0.278 & \textbf{0.269} \\
     & 720 & 0.353 & 0.395 & \textbf{0.340} & 0.361 & \textbf{0.340} & 0.348 & 0.363 & 0.395 & 0.594 & 0.375 & \textbf{0.349} & \textbf{0.349} \\
     & Avg & 0.246 & 0.350 & 0.243 & 0.251 & 0.241 & \textbf{0.238} & 0.259 & 0.308 & 0.456 & 0.278 & 0.259 & \textbf{0.253} \\
     \midrule  
    \multirow{5}{*}{\textbf{Electricity}} & 96 & 0.170 & 0.142 & 0.142 & 0.155 & 0.149 & \textbf{0.136} & \textbf{0.159} & 0.341 & 0.171 & 0.176 & 0.168 & \textbf{0.159} \\
     & 192 & 0.191 & 0.160 & 0.163 & 0.169 & 0.164 & \textbf{0.158} & 0.172 & 0.278 & 0.179 & 0.184 & 0.176 & \textbf{0.172} \\
     & 336 & 0.264 & 0.178 & 0.178 & 0.184 & 0.176 & \textbf{0.173} & 0.190 & 0.266 & 0.198 & 0.201 & 0.192 & \textbf{0.189} \\
     & 720 & 0.412 & 0.210 & 0.207 & 0.221 & \textbf{0.203} & 0.218 & 0.232 & 0.268 & 0.233 & 0.236 & \textbf{0.227} & 0.231 \\
     & Avg & 0.259 & 0.173 & 0.173 & 0.182 & 0.173 & \textbf{0.171} & 0.188 & 0.288 & 0.195 & 0.199 & 0.191 & \textbf{0.187} \\
     \midrule  
    \multirow{5}{*}{\textbf{Solar}} & 96 & 0.207 & 1.921 & 0.184 & \textbf{0.172} & 0.184 & 0.189 & \textbf{0.184} & 1.944 & 0.199 & 0.281 & 0.227 & 0.189 \\
     & 192 & 0.196 & 1.838 & \textbf{0.195} & 0.197 & \textbf{0.195} & 0.196 & 0.213 & 1.890 & 0.409 & 0.334 & 0.263 & \textbf{0.200} \\
     & 336 & 0.193 & 1.892 & 0.209 & 0.206 & 0.209 & \textbf{0.193} & 0.221 & 1.877 & 0.401 & 0.351 & 0.270 & \textbf{0.214} \\
     & 720 & 0.222 & 1.820 & 0.237 & 0.562 & 0.240 & \textbf{0.207} & 0.228 & 1.974 & 0.224 & 0.371 & 0.277 & \textbf{0.211} \\
     & Avg & 0.205 & 1.868 & 0.206 & 0.284 & 0.207 & \textbf{0.196} & 0.211 & 1.921 & 0.308 & 0.334 & 0.259 & \textbf{0.204} \\
     \midrule  
    \multirow{5}{*}{\textbf{AQshunyi}} & 96 & 0.704 & 1.234 & 0.580 & \textbf{0.542} & 0.635 & 0.647 & 0.765 & 1.239 & 0.668 & \textbf{0.663} & 0.683 & 0.669 \\
     & 192 & 0.691 & 0.883 & 0.637 & \textbf{0.580} & 0.668 & 0.702 & 0.799 & 1.002 & 0.703 & 0.703 & 0.707 & \textbf{0.702} \\
     & 336 & 0.723 & 0.862 & 0.682 & \textbf{0.612} & 0.704 & 0.727 & 0.837 & 0.888 & \textbf{0.721} & 0.726 & 0.727 & 0.723 \\
     & 720 & 0.782 & 0.811 & 0.712 & \textbf{0.663} & 0.735 & 0.776 & 0.881 & 0.810 & 0.758 & 0.769 & 0.757 & \textbf{0.754} \\
     & Avg & 0.725 & 0.948 & 0.653 & \textbf{0.599} & 0.686 & 0.713 & 0.821 & 0.985 & \textbf{0.712} & 0.715 & 0.718 & \textbf{0.712} \\
     \bottomrule
    \end{tabular}
    }
    }
\end{table}

\begin{table}[t] 
  \centering
  \renewcommand{\arraystretch}{1.1} 
  \caption{Full MAE results on UniTS and Timer.}
  \label{tab:full_mse_timer}
  \resizebox{1.0\textwidth}{!}{
  {
    \footnotesize
    \begin{tabular}{clllllllllllll}
    \toprule 
    & \multicolumn{1}{c}{} & \multicolumn{6}{c}{\textbf{Timer}} & \multicolumn{6}{c}{\textbf{UniTS}} \\
    \midrule  
    \multicolumn{1}{l}{} &  & \textit{Full} & \textit{LoRA} & \textit{AdaLoRA} & \textit{Linear} & \textit{MSFT} & \textit{Ours} & \textit{Full} & \textit{LoRA} & \textit{AdaLoRA} & \textit{Linear} & \textit{MSFT} & \textit{Ours} \\
    \midrule  
    \multirow{5}{*}{\textbf{ETTh1}} & 96 & 0.424 & \textbf{0.395} & 0.397 & \textbf{0.395} & 0.414 & 0.402 & 0.404 & 0.406 & 0.395 & 0.395 & \textbf{0.392} & 0.397 \\
     & 192 & 0.434 & 0.429 & 0.440 & 0.428 & 0.438 & \textbf{0.414} & 0.435 & 0.427 & 0.423 & 0.424 & \textbf{0.421} & 0.424 \\
     & 336 & 0.458 & 0.456 & 0.450 & 0.449 & 0.447 & \textbf{0.425} & 0.458 & \textbf{0.443} & 0.447 & 0.451 & 0.446 & 0.445 \\
     & 720 & 0.480 & 0.484 & 0.486 & 0.483 & 0.470 & \textbf{0.445} & 0.486 & 0.480 & 0.474 & 0.482 & \textbf{0.461} & 0.469 \\
     & Avg & 0.449 & 0.441 & 0.443 & 0.439 & 0.442 & \textbf{0.422} & 0.446 & 0.439 & 0.435 & 0.438 & \textbf{0.430} & 0.434 \\
     \midrule  
    \multirow{5}{*}{\textbf{ETTh2}} & 96 & 0.366 & 0.346 & 0.344 & 0.344 & 0.349 & \textbf{0.342} & 0.365 & 0.367 & 0.348 & 0.363 & 0.353 & \textbf{0.343} \\
     & 192 & 0.410 & 0.396 & 0.395 & 0.397 & 0.398 & \textbf{0.394} & 0.453 & 0.412 & 0.402 & 0.427 & 0.395 & \textbf{0.391} \\
     & 336 & 0.438 & 0.435 & 0.440 & 0.432 & 0.437 & \textbf{0.428} & 0.437 & 0.433 & 0.428 & 0.427 & 0.427 & \textbf{0.420} \\
     & 720 & 0.456 & 0.454 & 0.447 & 0.446 & 0.451 & \textbf{0.441} & 0.458 & 0.444 & 0.444 & 0.461 & 0.444 & \textbf{0.433} \\
     & Avg & 0.417 & 0.408 & 0.407 & 0.405 & 0.409 & \textbf{0.401} & 0.428 & 0.414 & 0.406 & 0.419 & 0.405 & \textbf{0.397} \\
     \midrule  
    \multirow{5}{*}{\textbf{ETTm1}} & 96 & 0.359 & 0.361 & 0.365 & 0.361 & 0.357 & \textbf{0.347} & 0.361 & 0.401 & 0.366 & 0.361 & 0.372 & \textbf{0.357} \\
     & 192 & 0.406 & 0.388 & 0.392 & 0.388 & 0.377 & \textbf{0.362} & 0.398 & 0.399 & 0.388 & 0.387 & 0.388 & \textbf{0.384} \\
     & 336 & 0.428 & 0.703 & 0.673 & 0.704 & 0.397 & \textbf{0.384} & 0.436 & 0.421 & 0.414 & 0.415 & 0.408 & \textbf{0.406} \\
     & 720 & 0.465 & 0.748 & 0.729 & 0.745 & 0.428 & \textbf{0.418} & 0.467 & 0.446 & 0.449 & 0.447 & 0.445 & \textbf{0.440} \\
     & Avg & 0.415 & 0.550 & 0.540 & 0.550 & 0.390 & \textbf{0.378} & 0.415 & 0.417 & 0.404 & 0.402 & 0.403 & \textbf{0.396} \\
     \midrule  
    \multirow{5}{*}{\textbf{ETTm2}} & 96 & 0.264 & 0.266 & 0.260 & 0.266 & 0.276 & \textbf{0.254} & 0.278 & 0.296 & 0.275 & 0.276 & 0.267 & \textbf{0.258} \\
     & 192 & 0.311 & 0.316 & 0.305 & 0.317 & 0.306 & \textbf{0.294} & 0.322 & 0.325 & 0.320 & 0.337 & 0.311 & \textbf{0.307} \\
     & 336 & 0.351 & 0.413 & 0.391 & 0.413 & 0.339 & \textbf{0.335} & 0.375 & 0.365 & 0.366 & 0.376 & 0.349 & \textbf{0.346} \\
     & 720 & 0.408 & 0.469 & 0.460 & 0.468 & \textbf{0.393} & 0.396 & 0.450 & 0.425 & 0.432 & 0.451 & \textbf{0.403} & 0.408 \\
     & Avg & 0.334 & 0.366 & 0.354 & 0.366 & 0.328 & \textbf{0.320} & 0.356 & 0.353 & 0.348 & 0.360 & 0.333 & \textbf{0.330} \\
     \midrule  
    \multirow{5}{*}{\textbf{Exchange}} & 96 & 0.233 & 0.208 & 0.207 & 0.202 & 0.215 & \textbf{0.199} & 0.234 & 0.276 & 0.211 & 0.214 & 0.212 & \textbf{0.204} \\
     & 192 & 0.349 & 0.302 & 0.304 & 0.298 & 0.379 & \textbf{0.294} & 0.325 & 0.345 & 0.309 & 0.320 & 0.307 & \textbf{0.304} \\
     & 336 & 0.419 & 0.422 & 0.422 & 0.411 & 0.488 & \textbf{0.406} & 0.437 & 0.444 & 0.426 & 0.421 & 0.427 & \textbf{0.412} \\
     & 720 & 0.819 & 0.703 & 0.706 & 0.699 & 0.770 & \textbf{0.689} & 0.712 & 0.695 & 0.689 & 0.688 & \textbf{0.680} & 0.687 \\
     & Avg & 0.455 & 0.409 & 0.410 & 0.402 & 0.463 & \textbf{0.397} & 0.427 & 0.440 & 0.409 & 0.411 & 0.407 & \textbf{0.402} \\
     \midrule  
    \multirow{5}{*}{\textbf{ZafNoo}} & 96 & 0.428 & 0.420 & 0.419 & 0.419 & 0.418 & \textbf{0.408} & 0.414 & 0.452 & \textbf{0.403} & \textbf{0.403} & 0.412 & 0.405 \\
     & 192 & 0.479 & 0.470 & 0.467 & 0.463 & 0.449 & \textbf{0.434} & 0.441 & 0.483 & 0.447 & 0.478 & 0.445 & \textbf{0.439} \\
     & 336 & 0.510 & 0.509 & 0.507 & 0.501 & 0.472 & \textbf{0.462} & 0.467 & 0.498 & 0.474 & 0.474 & \textbf{0.465} & 0.468 \\
     & 720 & 0.556 & 0.566 & 0.568 & 0.555 & 0.503 & \textbf{0.502} & 0.501 & 0.533 & 0.520 & 0.520 & 0.519 & \textbf{0.488} \\
     & Avg & 0.493 & 0.491 & 0.490 & 0.485 & 0.460 & \textbf{0.452} & 0.456 & 0.491 & 0.461 & 0.469 & 0.460 & \textbf{0.450} \\
     \midrule  
    \multirow{5}{*}{\textbf{Weather}} & 96 & 0.218 & 0.402 & 0.217 & 0.218 & 0.231 & \textbf{0.196} & 0.218 & 0.211 & 0.211 & 0.207 & 0.208 & \textbf{0.196} \\
     & 192 & 0.261 & 0.442 & 0.419 & 0.360 & 0.257 & \textbf{0.238} & 0.261 & 0.256 & 0.254 & 0.264 & 0.249 & \textbf{0.249} \\
     & 336 & 0.305 & 0.471 & 0.423 & 0.422 & 0.294 & \textbf{0.279} & 0.305 & 0.299 & 0.296 & 0.321 & \textbf{0.290} & 0.292 \\
     & 720 & 0.356 & 0.472 & 0.452 & 0.476 & 0.343 & \textbf{0.330} & 0.356 & 0.348 & 0.347 & 0.384 & \textbf{0.342} & 0.351 \\
     & Avg & 0.285 & 0.447 & 0.378 & 0.369 & 0.281 & \textbf{0.261} & 0.285 & 0.279 & 0.277 & 0.294 & 0.272 & \textbf{0.272} \\
     \midrule  
    \multirow{5}{*}{\textbf{Electricity}} & 96 & \textbf{0.224} & 0.410 & 0.252 & 0.276 & 0.253 & 0.239 & \textbf{0.231} & 0.259 & 0.243 & 0.264 & 0.252 & 0.239 \\
     & 192 & 0.244 & 0.411 & 0.266 & 0.295 & 0.266 & \textbf{0.243} & \textbf{0.249} & 0.266 & 0.259 & 0.280 & 0.266 & 0.260 \\
     & 336 & 0.262 & 0.410 & 0.366 & 0.318 & 0.281 & \textbf{0.261} & \textbf{0.273} & 0.285 & 0.275 & 0.294 & 0.282 & 0.276 \\
     & 720 & \textbf{0.297} & 0.417 & 0.401 & 0.364 & 0.313 & \textbf{0.297} & \textbf{0.295} & 0.316 & 0.307 & 0.326 & 0.315 & 0.297 \\
     & Avg & \textbf{0.257} & 0.412 & 0.321 & 0.313 & 0.278 & 0.260 & \textbf{0.262} & 0.282 & 0.271 & 0.291 & 0.279 & 0.268 \\
     \midrule  
    \multirow{5}{*}{\textbf{Solar}} & 96 & 0.245 & 0.272 & 0.310 & 0.310 & 0.262 & \textbf{0.225} & 0.219 & 0.968 & 0.334 & 0.227 & 0.241 & \textbf{0.213} \\
     & 192 & 0.252 & 0.327 & 0.405 & 0.341 & 0.270 & \textbf{0.237} & 0.230 & 0.991 & 0.332 & 0.232 & 0.249 & \textbf{0.225} \\
     & 336 & \textbf{0.244} & 0.327 & 0.413 & 0.366 & 0.279 & 0.248 & 0.238 & 1.018 & 0.331 & 0.242 & 0.260 & \textbf{0.235} \\
     & 720 & 0.282 & 0.328 & 0.428 & 0.397 & 0.288 & \textbf{0.264} & \textbf{0.240} & 0.994 & 0.478 & 0.248 & 0.273 & 0.250 \\
     & Avg & 0.256 & 0.314 & 0.389 & 0.354 & 0.275 & \textbf{0.244} & 0.232 & 0.993 & 0.369 & 0.237 & 0.256 & \textbf{0.231} \\
     \midrule  
    \multirow{5}{*}{\textbf{AQshunyi}} & 96 & 0.497 & 0.590 & 0.583 & 0.566 & 0.572 & \textbf{0.484} & 0.501 & 0.568 & 0.510 & 0.513 & 0.505 & \textbf{0.495} \\
     & 192 & 0.538 & 0.647 & 0.527 & \textbf{0.505} & 0.511 & 0.539 & 0.517 & 0.578 & 0.532 & 0.532 & 0.526 & \textbf{0.516} \\
     & 336 & 0.550 & 0.645 & 0.537 & \textbf{0.520} & 0.524 & 0.531 & 0.537 & 0.554 & 0.542 & 0.541 & 0.540 & \textbf{0.529} \\
     & 720 & 0.575 & 0.647 & 0.562 & 0.549 & 0.551 & \textbf{0.509} & 0.553 & 0.569 & 0.563 & 0.560 & 0.550 & \textbf{0.549} \\
     & Avg & 0.540 & 0.632 & 0.552 & 0.535 & 0.539 & \textbf{0.516} & 0.527 & 0.567 & 0.537 & 0.537 & 0.530 & \textbf{0.522} \\
     \bottomrule  
    \end{tabular}
    }
    }
\end{table}

\begin{table}[t] 
  \centering
  \renewcommand{\arraystretch}{1.1} 
  \caption{Full MAE results on Moirai and Sundial.}
  \label{tab:full_mse_timer}
  \resizebox{1.0\textwidth}{!}{
  {
    \footnotesize
    \begin{tabular}{clllllllllllll}
    \toprule 
     & \multicolumn{1}{c}{} & \multicolumn{6}{c}{\textbf{Moirai}} & \multicolumn{6}{c}{\textbf{Sundial}} \\
     \midrule  
    \multicolumn{1}{l}{} &  & \textit{Full} & \textit{LoRA} & \textit{AdaLoRA} & \textit{Linear} & \textit{MSFT} & \textit{Ours} & \textit{Full} & \textit{LoRA} & \textit{AdaLoRA} & \textit{Linear} & \textit{MSFT} & \textit{Ours} \\
    \midrule  
    \multirow{5}{*}{\textbf{ETTh1}} & 96 & 0.436 & 0.466 & \textbf{0.407} & 0.423 & 0.431 & \textbf{0.407} & 0.399 & \textbf{0.397} & 0.398 & 0.459 & 0.475 & \textbf{0.397} \\
     & 192 & 0.472 & 0.458 & 0.492 & 0.468 & 0.457 & \textbf{0.436} & 0.434 & 0.451 & 0.442 & 0.500 & 0.514 & \textbf{0.432} \\
     & 336 & 0.476 & 0.462 & 0.463 & 0.488 & \textbf{0.458} & 0.520 & 0.467 & 0.465 & 0.478 & 0.530 & 0.548 & \textbf{0.463} \\
     & 720 & 0.519 & 0.563 & 0.618 & 0.532 & \textbf{0.489} & 0.540 & 0.525 & \textbf{0.504} & 0.625 & 0.581 & 0.596 & 0.528 \\
     & Avg & 0.476 & 0.487 & 0.495 & 0.478 & \textbf{0.459} & 0.476 & 0.456 & \textbf{0.454} & 0.486 & 0.517 & 0.533 & 0.455 \\
     \midrule  
    \multirow{5}{*}{\textbf{ETTh2}} & 96 & 0.371 & 0.418 & 0.357 & 0.356 & 0.350 & \textbf{0.341} & 0.406 & 0.415 & 0.384 & 0.465 & 0.501 & \textbf{0.375} \\
     & 192 & 0.409 & 0.419 & 0.400 & 0.406 & 0.382 & \textbf{0.376} & 0.489 & \textbf{0.439} & 0.469 & 0.568 & 0.611 & 0.458 \\
     & 336 & 0.418 & 0.426 & 0.434 & 0.412 & 0.420 & \textbf{0.402} & 0.486 & \textbf{0.460} & 0.501 & 0.644 & 0.709 & 0.472 \\
     & 720 & 0.461 & 0.476 & \textbf{0.424} & 0.469 & 0.463 & 0.486 & 0.634 & \textbf{0.564} & 0.892 & 0.787 & 0.861 & 0.627 \\
     & Avg & 0.415 & 0.434 & 0.404 & 0.411 & 0.404 & \textbf{0.401} & 0.504 & \textbf{0.470} & 0.562 & 0.616 & 0.671 & 0.483 \\
     \midrule  
    \multirow{5}{*}{\textbf{ETTm1}} & 96 & \textbf{0.353} & 0.379 & 0.371 & 0.374 & 0.364 & 0.358 & 0.371 & 0.381 & 0.560 & 0.493 & 0.375 & \textbf{0.365} \\
     & 192 & 0.403 & 0.393 & 0.391 & 0.401 & \textbf{0.371} & 0.395 & 0.395 & 0.401 & 0.403 & 0.516 & 0.402 & \textbf{0.395} \\
     & 336 & 0.422 & 0.419 & 0.416 & 0.419 & 0.452 & \textbf{0.413} & 0.440 & \textbf{0.420} & 0.437 & 0.532 & 0.425 & 0.432 \\
     & 720 & 0.457 & 0.449 & 0.469 & 0.468 & 0.452 & \textbf{0.437} & 0.462 & 0.472 & 0.670 & 0.565 & \textbf{0.460} & 0.469 \\
     & Avg & 0.409 & 0.410 & 0.412 & 0.416 & 0.410 & \textbf{0.400} & 0.417 & 0.419 & 0.517 & 0.527 & \textbf{0.415} & \textbf{0.415} \\
     \midrule  
    \multirow{5}{*}{\textbf{ETTm2}} & 96 & 0.283 & 0.296 & 0.270 & 0.270 & 0.282 & \textbf{0.268} & 0.307 & 0.301 & 0.478 & 0.346 & 0.316 & \textbf{0.293} \\
     & 192 & 0.335 & 0.325 & 0.322 & 0.321 & 0.317 & \textbf{0.304} & 0.382 & \textbf{0.330} & 0.356 & 0.417 & 0.379 & 0.359 \\
     & 336 & 0.374 & 0.365 & 0.352 & \textbf{0.350} & 0.352 & 0.371 & 0.426 & \textbf{0.386} & 0.459 & 0.501 & 0.443 & 0.428 \\
     & 720 & 0.430 & 0.442 & 0.404 & 0.407 & 0.404 & \textbf{0.402} & 0.592 & 0.594 & 0.871 & 0.614 & \textbf{0.504} & 0.530 \\
     & Avg & 0.356 & 0.357 & 0.337 & 0.337 & 0.339 & \textbf{0.336} & 0.427 & 0.403 & 0.541 & 0.469 & 0.410 & \textbf{0.402} \\
     \midrule  
    \multirow{5}{*}{\textbf{Exchange}} & 96 & 0.213 & 0.300 & 0.246 & 0.245 & \textbf{0.198} & 0.204 & 0.252 & 0.298 & 0.292 & 0.257 & 0.281 & \textbf{0.239} \\
     & 192 & 0.314 & 0.353 & 0.353 & 0.406 & \textbf{0.301} & 0.302 & 0.369 & 0.341 & 0.369 & 0.377 & 0.410 & \textbf{0.330} \\
     & 336 & 0.439 & 0.478 & 0.508 & 0.510 & \textbf{0.419} & 0.487 & 0.531 & 0.510 & 0.522 & 0.502 & 0.538 & \textbf{0.448} \\
     & 720 & 0.778 & 0.767 & 0.903 & 0.780 & 0.726 & \textbf{0.689} & 0.769 & 0.804 & 0.757 & 0.746 & 0.747 & \textbf{0.680} \\
     & Avg & 0.436 & \multicolumn{1}{c}{0.474} & {0.502} & {0.485} & \textbf{0.411} & 0.421 & 0.480 & 0.488 & 0.485 & 0.470 & 0.494 & \textbf{0.424} \\
     \midrule  
    \multirow{5}{*}{\textbf{ZafNoo}} & 96 & 0.427 & 0.514 & 0.431 & 0.433 & 0.395 & \textbf{0.392} & 0.416 & 0.511 & 0.448 & 0.421 & 0.434 & \textbf{0.391} \\
     & 192 & 0.465 & 0.493 & 0.463 & 0.455 & \textbf{0.418} & 0.436 & 0.443 & 0.489 & 0.448 & 0.460 & 0.476 & \textbf{0.428} \\
     & 336 & 0.470 & 0.497 & 0.477 & 0.476 & \textbf{0.467} & 0.469 & 0.452 & 0.480 & 0.456 & 0.483 & 0.501 & \textbf{0.444} \\
     & 720 & 0.510 & 0.527 & 0.470 & 0.509 & \textbf{0.458} & 0.466 & \textbf{0.467} & 0.479 & 0.554 & 0.513 & 0.531 & 0.468 \\
     & Avg & 0.468 & 0.508 & 0.460 & 0.468 & \textbf{0.435} & 0.441 & 0.445 & 0.490 & 0.476 & 0.469 & 0.485 & \textbf{0.433} \\
     \midrule  
    \multirow{5}{*}{\textbf{Weather}} & 96 & 0.204 & 0.331 & 0.204 & 0.202 & \textbf{0.195} & 0.200 & 0.219 & 0.308 & 0.314 & 0.257 & 0.230 & \textbf{0.213} \\
     & 192 & \textbf{0.241} & 0.339 & 0.255 & 0.252 & 0.255 & 0.242 & 0.269 & 0.311 & 0.382 & 0.302 & 0.278 & \textbf{0.255} \\
     & 336 & 0.295 & 0.367 & 0.287 & 0.293 & 0.300 & \textbf{0.279} & 0.314 & 0.340 & 0.422 & 0.341 & 0.316 & \textbf{0.297} \\
     & 720 & 0.342 & 0.390 & \textbf{0.339} & 0.354 & \textbf{0.339} & 0.360 & 0.367 & 0.388 & 0.494 & 0.398 & 0.370 & \textbf{0.358} \\
     & Avg & 0.271 & 0.357 & 0.271 & 0.275 & 0.272 & \textbf{0.270} & 0.292 & 0.337 & 0.403 & 0.324 & 0.298 & \textbf{0.281} \\
     \midrule  
    \multirow{5}{*}{\textbf{Electricity}} & 96 & 0.258 & 0.234 & 0.235 & 0.246 & 0.248 & \textbf{0.234} & 0.247 & 0.350 & 0.258 & 0.265 & 0.256 & \textbf{0.246} \\
     & 192 & 0.278 & \textbf{0.252} & 0.256 & 0.258 & 0.262 & 0.254 & 0.260 & 0.325 & 0.268 & 0.274 & 0.266 & \textbf{0.260} \\
     & 336 & 0.334 & 0.269 & \textbf{0.267} & 0.272 & 0.278 & 0.272 & 0.282 & 0.326 & 0.285 & 0.291 & 0.284 & \textbf{0.280} \\
     & 720 & 0.337 & 0.296 & 0.296 & 0.301 & \textbf{0.293} & 0.309 & \textbf{0.314} & 0.339 & 0.317 & 0.325 & 0.316 & \textbf{0.314} \\
     & Avg & 0.302 & \textbf{0.263} & 0.264 & 0.269 & 0.270 & 0.267 & 0.276 & 0.335 & 0.282 & 0.289 & 0.280 & \textbf{0.275} \\
     \midrule  
    \multirow{5}{*}{\textbf{Solar}} & 96 & 0.222 & 0.990 & 0.243 & \textbf{0.208} & 0.216 & 0.216 & 0.239 & 1.015 & 0.253 & 0.337 & 0.277 & \textbf{0.230} \\
     & 192 & 0.219 & 0.980 & 0.247 & 0.209 & \textbf{0.205} & \textbf{0.205} & 0.263 & 1.032 & 0.375 & 0.375 & 0.305 & \textbf{0.252} \\
     & 336 & 0.209 & 1.004 & 0.241 & 0.226 & 0.219 & \textbf{0.219} & 0.272 & 1.081 & 0.371 & 0.385 & 0.310 & \textbf{0.254} \\
     & 720 & 0.234 & 0.987 & 0.237 & 0.577 & 0.253 & \textbf{0.229} & 0.275 & 1.092 & 0.274 & 0.340 & 0.321 & \textbf{0.254} \\
     & Avg & 0.221 & 0.990 & 0.242 & 0.305 & 0.223 & \textbf{0.217} & 0.262 & 1.055 & 0.318 & 0.359 & 0.303 & \textbf{0.247} \\
     \midrule  
    \multirow{5}{*}{\textbf{AQshunyi}} & 96 & 0.495 & 0.655 & \textbf{0.431} & 0.437 & 0.451 & 0.483 & 0.527 & 0.661 & 0.491 & 0.501 & 0.501 & \textbf{0.489} \\
     & 192 & 0.502 & 0.565 & \textbf{0.460} & \textbf{0.460} & 0.468 & 0.501 & 0.543 & 0.602 & 0.510 & 0.512 & 0.510 & \textbf{0.507} \\
     & 336 & 0.514 & 0.561 & \textbf{0.482} & 0.491 & 0.492 & 0.518 & 0.557 & 0.574 & 0.524 & 0.534 & \textbf{0.519} & 0.523 \\
     & 720 & 0.539 & 0.545 & 0.492 & \textbf{0.491} & 0.504 & 0.536 & 0.575 & 0.551 & 0.542 & 0.541 & 0.542 & \textbf{0.534} \\
     & Avg & 0.512 & 0.581 & \textbf{0.466} & 0.470 & 0.479 & 0.510 & 0.551 & 0.597 & 0.517 & 0.522 & 0.518 & \textbf{0.513} \\
    \bottomrule  
    \end{tabular}
    }
    }
\end{table}

\end{document}

%% file: math_commands.tex
\usepackage{amsmath,amsfonts,bm}

\def\eqref#1{equation~\ref{#1}}

\def\1{\bm{1}}

\DeclareMathAlphabet{\mathsfit}{\encodingdefault}{\sfdefault}{m}{sl}
\SetMathAlphabet{\mathsfit}{bold}{\encodingdefault}{\sfdefault}{bx}{n}

